\documentclass[journal]{IEEEtran}

\usepackage[pdftex]{graphicx}
\usepackage{amsmath}
\usepackage{array}
\usepackage{url}
\usepackage{graphicx}
\usepackage{amsmath,amssymb} 
\usepackage{color}
\usepackage[utf8]{inputenc}
\usepackage{version}
\usepackage[tight, footnotesize]{subfigure}
\usepackage{relsize}
\usepackage{paralist}
\usepackage[super]{nth}
\usepackage{capt-of,etoolbox}
\usepackage{multirow}
\usepackage{xcolor}
\usepackage{algorithmic}
\usepackage[textwidth=3.6cm, textsize=footnotesize]{todonotes}
\DeclareMathOperator{\argmin}{argmin}

\DeclareMathOperator{\argmax}{argmax}

\newcommand{\LL}{\mathcal{L}}
\newcommand{\KK}{\mathcal{K}}

\newcommand{\thetaiiprim}{\theta _{ii'}}

\newcommand{\spatch}[1]{\ensuremath{\mathbf{#1}}}
\definecolor{gg}{RGB}{0,100,0}
\newcommand{\eqsize}{\normalsize}
\newcommand{\sss}{\scriptsize}

\usepackage[flushleft]{threeparttable}

\begin{document}

\title{SuperPatchMatch: an Algorithm for Robust Correspondences using Superpixel Patches}

\author{Rémi~Giraud, 
        Vinh-Thong~Ta, 
        Aurélie~Bugeau,
	Pierrick~Coupé,
        and~Nicolas~Papadakis 
        \thanks{This study has been carried out with financial support from the French 
State, managed by the French National Research Agency (ANR) in the  
frame of the GOTMI project (ANR-16-CE33-0010-01) and
the Investments for the future Program IdEx Bordeaux
(ANR-10-IDEX-03-02) with the Cluster of excellence CPU and TRAIL (HR-DTI
ANR-10-LABX-57).}
\thanks{R. Giraud is with LaBRI, UMR 5800 and IMB, UMR 5251, F-33400 Talence, France. 
email: remi.giraud@labri.fr. phone: +33(0) 5 4000 6937.}
\thanks{
V.-T. Ta is with Bordeaux INP, LaBRI, UMR 5800, F-33400 Talence, France.
email: vinh-thong.ta@labri.fr. phone: +33(0) 5 4000 3538.}
\thanks{
A. Bugeau is with University of Bordeaux, LaBRI, UMR 5800, F-33400 Talence, France.
email: aurelie.bugeau@labri.fr. phone: +33(0) 5 4000 3528.}
\thanks{
P. Coupé is with CNRS, LaBRI, UMR 5800, F-33400 Talence, France.
email: pierrick.coupe@labri.fr. phone: +33(0) 5 4000 3538.}
\thanks{N. Papadakis is with CNRS, IMB, UMR 5251, F-33400 Talence, France.
email: nicolas.papadakis@math.u-bordeaux.fr. phone: +33(0) 5 4000 2116.
}
}
        \maketitle
\IEEEpeerreviewmaketitle

\begin{abstract}

Superpixels have become very popular 
in many computer vision applications.
Nevertheless, they remain underexploited since 
the superpixel decomposition may produce 
irregular and non stable segmentation results 
due to the dependency to the image content.
In this paper,  we first
introduce a novel structure, a superpixel-based patch, called SuperPatch.  
The proposed structure, based on superpixel neighborhood, leads to a robust descriptor since spatial
information is naturally included.
The generalization of the  PatchMatch method to
SuperPatches,  
named SuperPatchMatch, is introduced. 
Finally, 
we propose a framework to perform fast segmentation and labeling from an image database, and
demonstrate the potential of our approach
since we 
outperform, in terms of computational cost and accuracy, the results of
state-of-the-art methods
on both face labeling and medical image segmentation.

\end{abstract}


\section{Introduction}

Image segmentation is a useful tool to analyze the image content. 
{\color{black}
The goal of segmentation is to decompose the image  into meaningful segments, 
for instance, to separate objects from the background.}
A segmentation is computed with respect to some priors such as
shape, color or texture.   
To reduce the  
computational cost, \emph{superpixel} decomposition  
methods have been developed for grouping pixels into
homogeneous regions, while respecting the image contours
(for instance see  \cite{achanta2012} and references therein). 
Superpixels are able to drastically decrease the number of elements to process 
while keeping all the geometrical information 
that is lost with multi-resolution approaches. 
Small objects disappear at low resolution levels, 
whereas they can still be represented with one or several superpixels.
Nevertheless, superpixels remain underexploited due to 
their irregular decomposition of the image content.

Many image processing and computer vision methods
use reference images.
For instance, for labeling applications, these images can be provided with their 
ground truth segmentation, labels, or semantic information
that are used to process the input image.
In this context, matching algorithms can be useful to find associations between the considered elements. 
In most frameworks, patch-based approximate nearest neighbor (ANN) search methods are used to find correspondences.
Numerous methods have been proposed to find ANN \cite{barnes2009,muja2009fast,korman2011,olonetsky2012} 
within the same image, and
between an image and one or several reference ones.
Among these methods, the  PatchMatch (PM) method \cite{barnes2009} was designed to compute correspondences between pixel-based patches.

When applying PM to large images, or when looking for ANN in a database, the search for good ANN may require many iterations. 
Therefore, multi-resolution  PM \cite{barnes2010}  can be considered to initialize the ANN correspondence map.
However, as usually observed with such coarse-to-fine frameworks, details are lost and a poor ANN is estimated for small scale patterns. 
A regular decomposition of the image could decrease the problem dimension, but it would not respect the object contours, leading to non accurate processing.
In this context, the use of superpixels may be interesting to
preserve the image geometry and the respect of the image object contours.
Local superpixel-based matching models have been proposed for many applications,
\emph{e.g.}, video tracking \cite{wang2011,reso2013}.
However, superpixel-based ANN search algorithms have been little investigated in the literature,
and recent works such as \cite{Rabin_icip14,liu2016photo}
that 
compute superpixel correspondences between the decompositions of two images,
use complex models that
report prohibitive computational times.

Finally, for ANN matching,
the neighborhood information greatly helps in finding good correspondences, as demonstrated
in the patch-based literature.
Therefore, to jointly decrease the number of elements to process,
keep the geometrical information,
and find accurate matches, it appears necessary to
consider superpixels 
and to describe them using their neighborhoods
in a structure that includes spatial information.
Nevertheless, the lack of regularity between two superpixel decompositions makes difficult
the use of neighborhood for computing relevant correspondences.
Some attempts to use superpixel neighborhood information have been proposed
\cite{pei2014,sawhney2014}. 
However, these methods are not adapted to the search of 
ANN, since they perform a regularization on a graph built from superpixel neighbors 
but do not include the relative 
spatial information between superpixels in a dedicated structure.

\newcommand\sih{0.07\textwidth}
\newcommand\sihh{0.115\textwidth}
\begin{figure*}[t!]
\centering
{\footnotesize
\begin{tabular}{@{\hspace{0mm}}c@{\hspace{3mm}}c@{\hspace{1mm}}c@{\hspace{1mm}}c@{\hspace{2mm}}}
\includegraphics[width=\sihh]{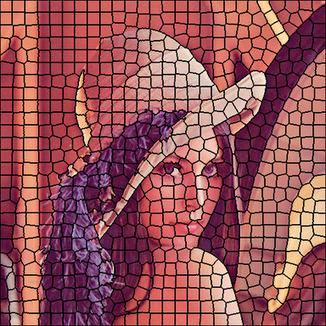}&
\includegraphics[width=\sihh]{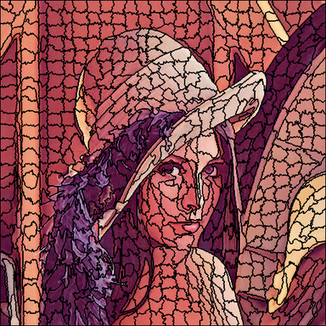}&
\includegraphics[width=\sihh]{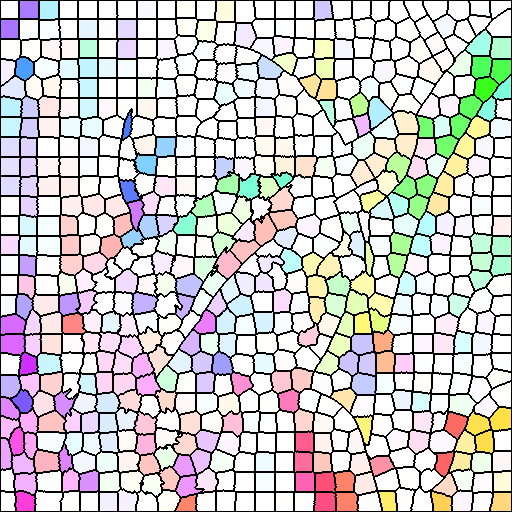}&
\includegraphics[width=\sihh]{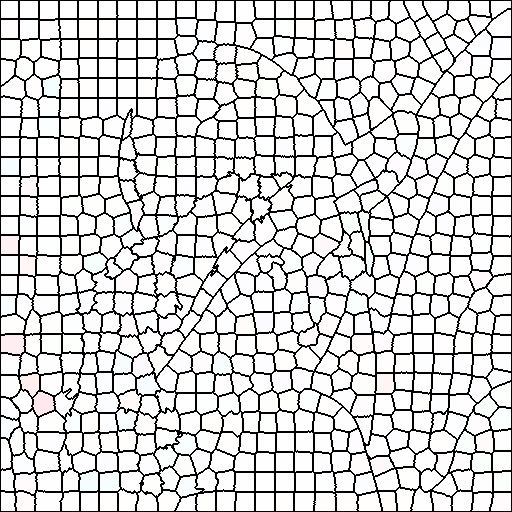}\\
(a)&(b)&(c)&(d)
\end{tabular}
\begin{tabular}{@{\hspace{0mm}}c@{\hspace{1mm}}c@{\hspace{1mm}}c@{\hspace{3mm}}c@{\hspace{0mm}}}
\includegraphics[width=\sihh,height=\sihh]{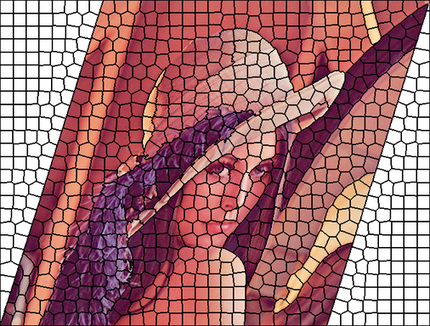}&
\includegraphics[width=\sihh]{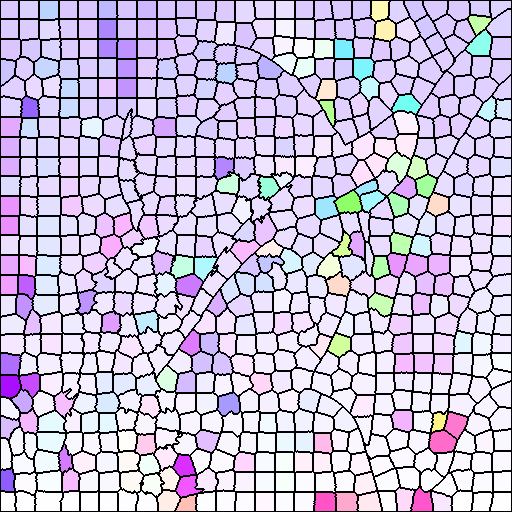}&
\includegraphics[width=\sihh]{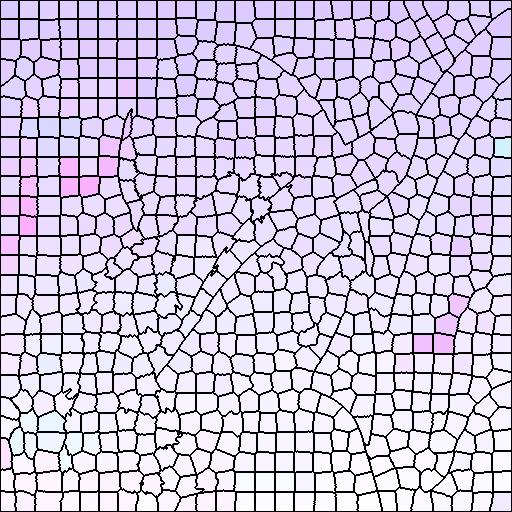}&
\includegraphics[width=\sihh,height=\sihh]{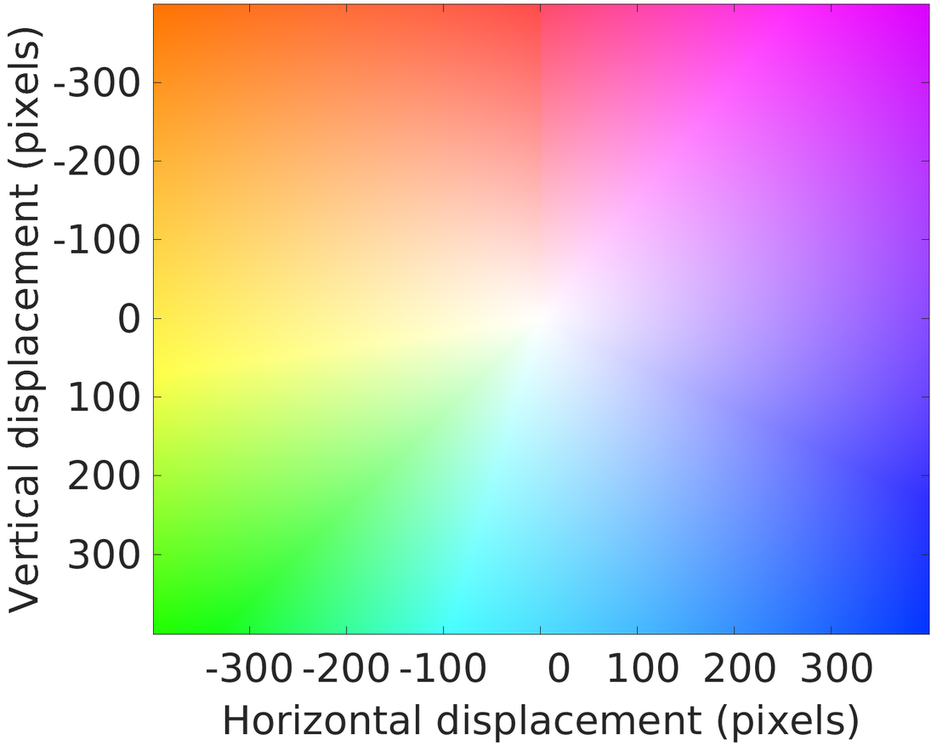}\\ 
(e)&(f)&(g)&(h)
\end{tabular}
}
\caption{Superpixels vs superpatches for superpixel matching.
(a) and (b): two decompositions using \cite{achanta2012} and \cite{buyssens2014}.
(c) and (d): superpixel-based \cite{gould2014} and our superpatch-based matching results. 
The same experiment is performed between (a) and the sheared image (e),
with superpixel \cite{gould2014} (f) and superpatch matching results (g).
The displacement is illustrated with optical flow representation (h).
The more the colored result is close to white, the lower the displacement is. 
}
\label{fig:sp_vector}
\end{figure*}

\begin{figure*}[t!]
\centering
{\footnotesize
\begin{tabular}{@{\hspace{0mm}}c@{\hspace{1mm}}c@{\hspace{2mm}}c@{\hspace{1mm}}c@{\hspace{5mm}}}
\includegraphics[width=\sihh]{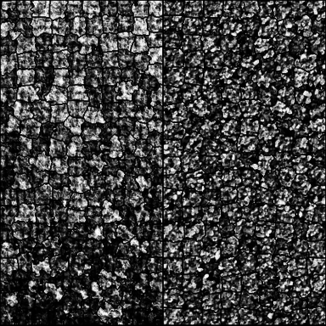}&
\includegraphics[width=\sihh]{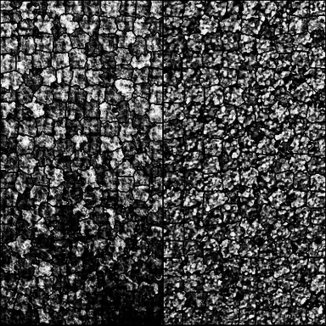}&
\includegraphics[width=\sihh]{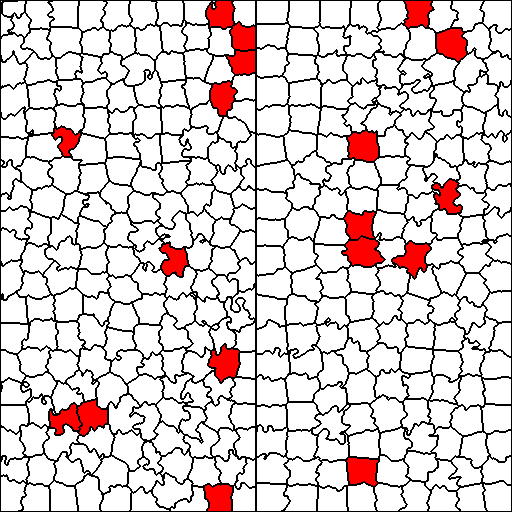}&
\includegraphics[width=\sihh]{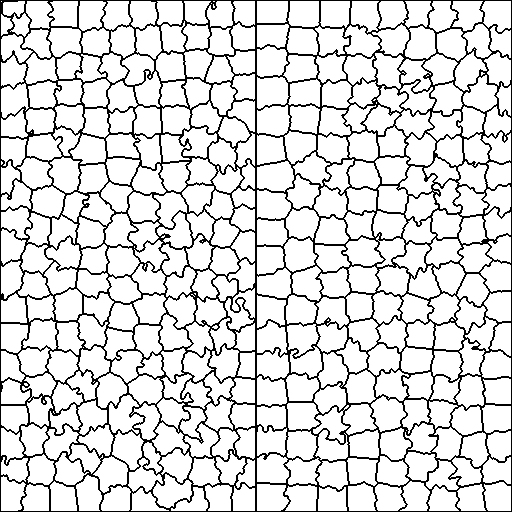}\\
(a)&(b)&(c)&(d)\\
\end{tabular}
\begin{tabular}{@{\hspace{0mm}}c@{\hspace{1mm}}c@{\hspace{2mm}}c@{\hspace{1mm}}c@{\hspace{0mm}}}
\includegraphics[width=\sihh]{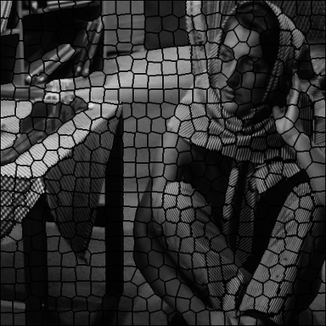}&
\includegraphics[width=\sihh]{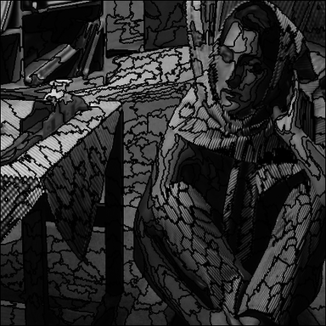}&
\includegraphics[width=\sihh]{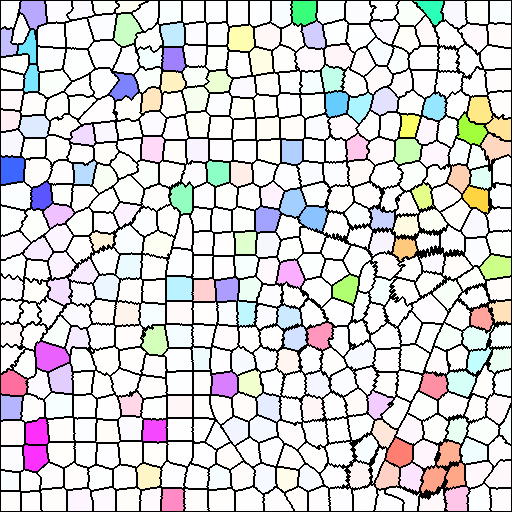}&
\includegraphics[width=\sihh]{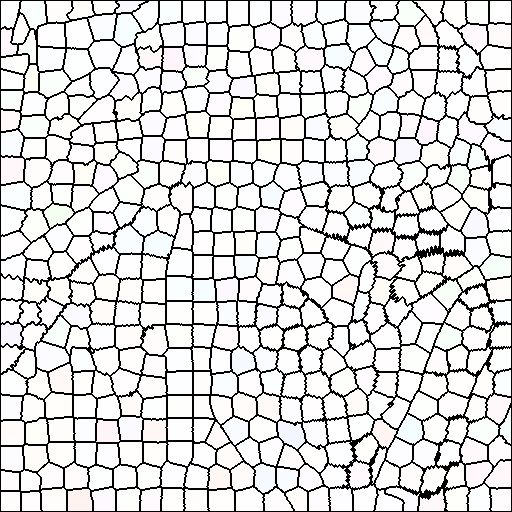}\\
(e)&(f)&(g)&(h)\\
\end{tabular}
}
\caption{Superpixel matching on textured images.
Two different parts of two close textures are combined in (a) and (b).
(c) and (d): superpixel-based \cite{gould2014}  and our superpatch-based matching results where
red superpixels indicate wrong matched texture.
(e) and (f): two decompositions of a natural textured image using \cite{achanta2012} and \cite{buyssens2014}.
Comparison of superpatch matching with color (g) and combination of color and texture features (h).
}
\label{fig:sp_vector2}
\end{figure*}

\subsection{Contributions} 
In this paper, we propose a novel structure of superpixel
neighborhood  called \emph{SuperPatch}. 
Since the superpixel neighborhoods of 
two superpatches are not necessarily the same (in terms of shape or  number of elements),
 a generic framework for comparing superpatches is introduced. 
A novel method, called {\em SuperPatchMatch} (SPM), that generalizes the PM algorithm \cite{barnes2009}, 
is proposed to perform fast and accurate searches of ANN superpatches within images.

To the best of our knowledge, the specific combination 
of PM with superpixels has been proposed in \cite{gould2014,zheng2015}
that propose to match single superpixels using moves similar to PM.
For instance in \cite{gould2014}, 
the superpixel features are pre-computed using a learned
distance metric,
while the reported labeling results do not reach the ones of state-of-the-art methods.
In \cite{lu2013}, a more restricted framework is considered for optical flow estimation:  PM is used to refine the results
within selected superpixel bounding boxes. The purpose of our work is thus completely different since we
compare neighborhood of structures defined on
irregular image sub-domains.

To emphasize the interest  of our method, 
we propose a framework to perform fast segmentation and labeling from an
image database.
SuperPatchMatch is
well adapted to deal with huge and constantly
growing databases since no learning phase is
required, contrary to most existing approaches
based on 
supervised machine learning \cite{he2004,kae2013},
or recent neural network methods \cite{liu2015,long2015}.
We apply  SuperPatchMatch to the challenging Labeled Faces in the Wild (LFW) 
database~\cite{huang2007db}, where the goal is to extract  hair,
 face, and  background within images decomposed into superpixels,
and to the segmentation of tumors on non-registered Magnetic Resonance Images (MRI).
Finally, SuperPatchMatch outperforms, in terms of computational cost and accuracy, 
state-of-the-art methods.

Fig. \ref{fig:sp_vector} and \ref{fig:sp_vector2} consider several experiments to demonstrate
that superpatches enable to find more reliable superpixel ANN 
than the ones obtained with single superpixel matching \cite{gould2014}.
In Fig. \ref{fig:sp_vector}, 
two decompositions are computed on the same image using \cite{achanta2012} 
and \cite{buyssens2014} (Fig. \ref{fig:sp_vector}(a) and (b)). 
The aim is to find the best superpixel match between (a) and (b)
in terms of superpixel feature (here $\ell_2$-norm on normalized color histograms in RGB space). 
We display the displacement magnitude of matches with optical flow representation (Fig. \ref{fig:sp_vector}(h)).
When matching only superpixels, as in \cite{gould2014},
many outliers are obtained (Fig. \ref{fig:sp_vector}(c)), 
while the matching of superpatches
provides very accurate ANN (Fig. \ref{fig:sp_vector}(d)).
The same experiment on a sheared image decomposed with \cite{achanta2012} (Fig. \ref{fig:sp_vector}(e)), 
provides a uniform displacement (Fig. \ref{fig:sp_vector}(g)) that indicates relevant superpatch matching,
and robustness of the proposed structure to geometrical deformations.
In Fig. \ref{fig:sp_vector2},
two different parts of two close textures are combined in Fig. \ref{fig:sp_vector2}(a) and (b),
and we represent  wrong matched texture with red superpixels in Fig. \ref{fig:sp_vector2}(c) and (d).
Finally, in Fig. \ref{fig:sp_vector2}, we show that for a natural image containing texture (Fig. \ref{fig:sp_vector2}(e) and (f)), 
the combination of color and texture features (histogram of oriented gradients \cite{dalal2005})
can provide more accurate matching (Fig. \ref{fig:sp_vector2}(h)).

\subsection{Outline} 
In this paper, 
we first present related works in Section \ref{sec:rw}.
Then, we define the new superpatch structure and 
a comparison framework between superpatches in Section \ref{section:SuperPatch}.
The SuperPatchMatch algorithm is next designed to perform superpixel-based ANN search in Section \ref{section:superpatchmatch}.
We further emphasize the interest of our method by proposing in Section \ref{section:expe}
a framework to perform labeling from an image database. 
Finally, we present experiments of face labeling and segmentation of medical images, and
 SuperPatchMatch results outperform  the ones of state-of-the-art methods.

\section{\label{sec:rw}Related Works}

\subsection{Superpixel Methods}
Superpixel decomposition approaches
try to group the pixels of an image into meaningful homogeneous regions.
They were progressively introduced, for instance, from watershed \cite{vincent1991} to
Quick shift \cite{vedaldi2008} approaches.
In the past years, most decomposition methods 
start from an initial regular grid
and refine the superpixel boundaries by computing a trade-off between 
color distance
and superpixel shape regularity, \emph{e.g.}, \cite{achanta2012,li2015}.
Recently, works such as \cite{machairas2015,giraud2016scalp} propose 
to use gradient and contour information in the process
to further increase the superpixel decomposition accuracy 
with respect to the image content.
Finally, the computational cost is considered
since superpixels are mainly used as pre-processing, and
recent implementations report real-time performances, \emph{e.g.}, \cite{ban2016}.

By considering features at the superpixel scale, the computational
complexity of computer vision and image processing tasks can be drastically reduced,
while still considering the image geometry and content. 
Superpixels have therefore become key building blocks
of many recent image processing and computer vision 
pipelines such as
multi-class object segmentation
\cite{gould2008,tighe2010,yang2010,mostajabi2015},  
body model estimation \cite{mori2005}, 
face and hair labeling \cite{kae2013}, 
data associations across views \cite{sawhney2014}, 
object localization \cite{fulkerson2009} or
contour detection \cite{arbelaez2011}.
With these considerations, we propose in this work to use the superpixel representation
as the basis of our framework.

\subsection{Including Spatial Information within Image Features}

Pixel-based patches enable to describe the pixel neighborhood and to find similar patterns
with the same geometric structure.
They have progressively proven their efficiency for several applications such as 
texture synthesis \cite{Efros99} and image denoising \cite{buades2005}, 
and in the design of computer vision descriptors \cite{Lowe2004,bay2006}
that include spatial information.

Recent works in object retrieval have demonstrated that 
describing the objects with spatial information enables to
reach higher detection accuracy. 
In \cite{garnier2012,clement2015}, Force-Histogram Decomposition descriptors are used to
encode the pairwise spatial relations between objects.
Deformable part models \cite{lsvm-pami,trulls2014} or adaptive
bounding boxes of poselets \cite{sharma2013} 
 have also been successfully applied to
image retrieval, segmentation or recognition. 
Finally, the necessity for including spatial information is also studied in \cite{bloch2005}
that investigates fuzzy approaches to define spatial relationships.

The superpixel itself is not sufficient to provide a
robust image descriptor,  
since the consistency of its neighborhood is not considered.
The superpixel neighborhood has been used in
\cite{pei2014} for saliency detection based on energy minimization.
For each superpixel, the two first adjacent neighbor rings are used in a
regularization term. However, the superpixel features are separately included
in a data term, leading to a lack of spatial information
consistency.
The approach is thus dependent on the superpixel decomposition and poorly robust to very irregular decomposition.
Consequently, we propose to go further in this work and to take
advantage of the superpixel neighborhood to construct a novel representation, namely
the superpatch, that naturally includes spatial information.

\subsection{Patch Matching Methods}
Patch-based methods have demonstrated state-of-the-art results over
various computer vision and image processing 
applications such as: texture synthesis
\cite{Efros99}, denoising 
\cite{buades2005} or super-resolution \cite{Freeman02}.
These approaches rely on the search of ANN, \emph{i.e.}, similar patches.
Many methods were proposed to find ANN
within
the image itself, between two images
or in an entire database \cite{barnes2009,muja2009fast,korman2011,olonetsky2012}.
When facing  huge databases,
dimension reduction methods 
are usually considered  to have fast
computation of ANN, but they depend on the size of the data.
In this context, the PatchMatch (PM) algorithm \cite{barnes2009} is an efficient
tool to compute ANN.  
Within an image itself,
the found ANN enable to perform several processings such as
image retargeting or completion \cite{barnes2009}.
Nevertheless, PM can also find matches between several images,
and easily handles large databases, since its complexity only depends
on the size of the image to process,
as shown in \cite{shi2013,giraud2016} 
where the ANN are used for exemplar-based segmentation of 3D medical images.

In this work, we introduce the SuperPatchMatch method (SPM), that
combines both the advantages of the PM algorithm, and the superpixel
decomposition of an image, to compute robust correspondences of superpixels
using superpatches. 
The proposed superpatch structure enables to match
similar patterns at the superpixel level since it considers
the geometrical information between the contained superpixels,
which are described by image features such as color or texture.

\section{\label{section:SuperPatch}Superpatch}

\subsection{\label{subsection:SuperPatch}Superpatch Definition}

\begin{figure}[t]
\centering
\newcommand{\siz}{0.27\textwidth}
\includegraphics[height=\siz]{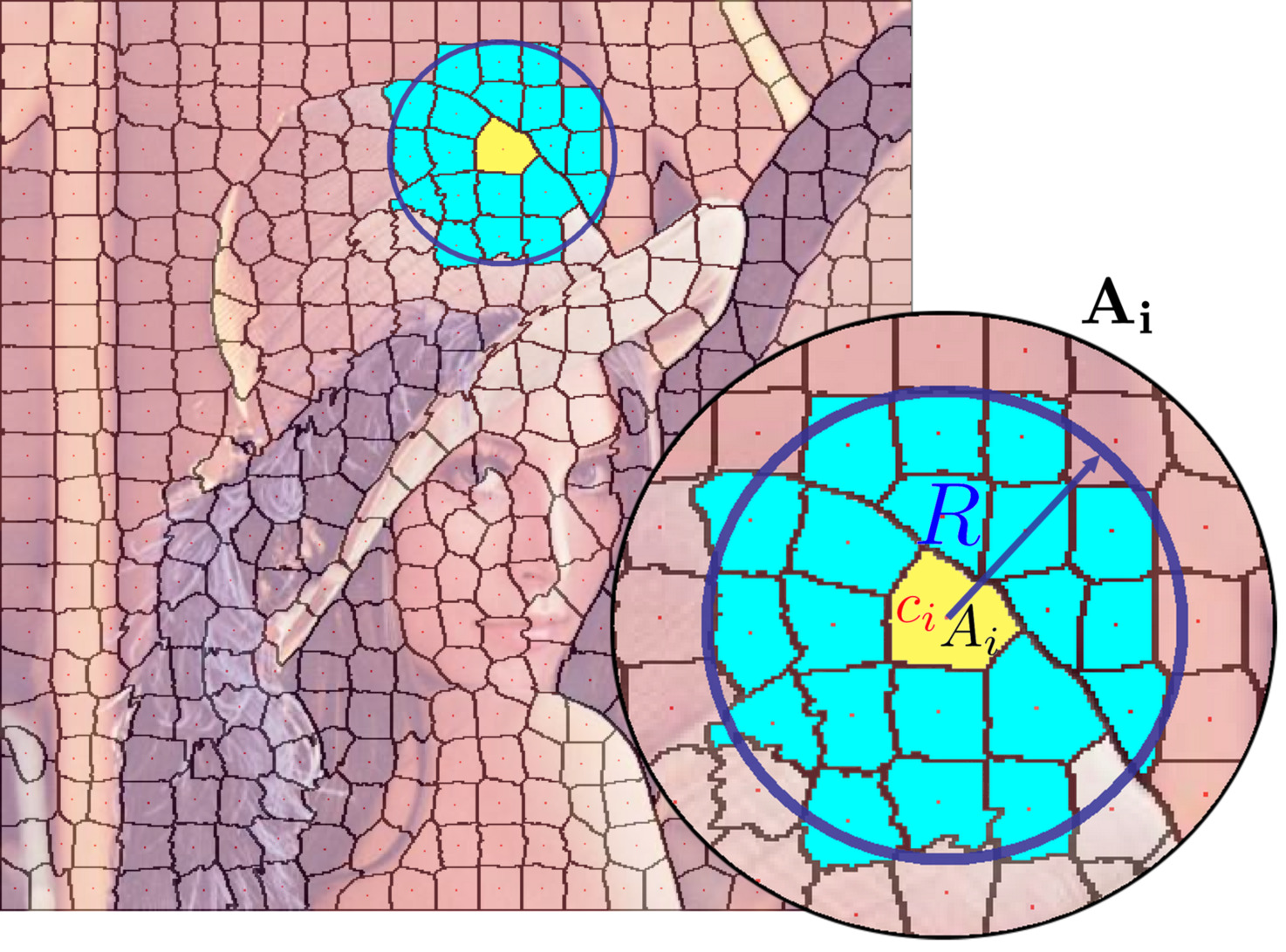}
\caption{Superpatch illustration. In blue: circle search of radius $R$ centered
on $c_i$, barycenter of $A_i$ (yellow). The superpatch $\spatch{A_i}$ is composed
by all  superpixels having their barycenter within the circle.
}
\label{fig:SP_seg}
\end{figure}

Similarly to a patch of pixels,   a \emph{superpatch} is  a patch
(a set) of neighboring  superpixels.
Let $A$ be an image, decomposed 
by any superpixel decomposition method,
into $|A|$ 
superpixels such that
$A=\{A_i\}_{i\in \{1,\dots,|A|\}}$,
where $|.|$ denotes the cardinality,
and for two superpixels $A_i,A_j\in A$, $A_i\cap A_j=\emptyset$.
A superpatch $\spatch{A_i}$ is centered on a superpixel $A_i$ and is
composed of its neighboring superpixels $A_{i'}$ such that:
$\spatch{A_i}=\{A_{i'}, \textrm{ with }  ||c_i-c_{i'}||\leq R\}$,
with $c_i$ the spatial barycenter of the pixels contained in $A_i$.
In other words, the superpatch centered on a superpixel is defined by
considering all superpixels within a fixed radius $R\geq 0$.
Note that each superpatch $\spatch{A_i}$
contains at least the superpixel $A_i$.
Fig. \ref{fig:SP_seg} illustrates the superpatch definition.

For the sake of clarity, we denote 
$\mathcal{I}^A_i$=$\{i', \textrm{ with }\,  A_{i'}$ $\in$ $\spatch{A_i}\}$, 
the index set of superpixels $A_{i'}\in \spatch{A_i}$.
Each superpixel $A_i$ is described by a set of features $F_i^A$.
These features can be, for instance, the coordinates of
$c_i$, the mean color,
or any superpixel descriptors that can be found in the literature.

\subsection{\label{subsection:SuperPatchComparison}Superpatch Comparison Framework}

The comparison between two regular square patches is commonly
performed using the sum of squared differences (SSD), 
computed
in a scan order.
When considering two superpatches, 
their number of elements and geometry are generally different,
which makes difficult their comparison.
In the following, we consider two superpatches
$\spatch{A_i}$ and
$\spatch{B_j}$, in different images $A$ and $B$.
We propose to first register the relative positions of all superpixels within the superpatches.
To overlap two superpatches, all positions $c_{j'}$ of
superpixels $B_{j'}\in \spatch{B_j}$ are registered with the vector
$v_{ij}=c_i-c_j$,  
where $c_i$ and $c_j$ are the spatial barycenters of ${A_i}$ and
${B_j}$, respectively.
Contrarily to the classical pixel setting, the number of 
elements and geometry of two superpatches are likely to
differ since their construction
depends on the initial
superpixel decomposition.
Therefore, a registered superpixel
$B_{j'}$
can overlap with several superpixels $A_{i'}$,
and this information has to be considered.

To compute a distance between irregular structures, such as superpixels, 
\cite{sawhney2014} proposes to use the editing distance. 
However, such distance computes one-to-one matching between the
structure elements and cannot accurately deal with the overlap
of superpixels that requires a one-to-many mapping.
Another limitation
is that it mixes two different information:  
superpixel similarities and the cost of removing or adding superpixels.  
Therefore, this distance should be carefully tuned with respect to the
considered application.   
Consequently, to define a relevant metric between
superpatches,  
it is necessary to consider the
geometry of the superpatches within the distance. 
We propose to define the symmetric distance $D$ between two superpatches
$\spatch{A_i}$ and $\spatch{B_j}$ as:  
{\eqsize
\begin{equation}
D(\spatch{A_i},\spatch{B_j})=\frac{\sum_{i'\in \mathcal{I}^A_i}\sum_{j'\in \mathcal{I}^B_j}^{} w(A_{i'},B_{j'}) d(F_{i'}^A,F_{j'}^B)}{\sum_{i'\in \mathcal{I}^A_i}\sum_{j'\in \mathcal{I}^B_j}^{} w(A_{i'},B_{j'})},
\label{D}
\end{equation}
}\noindent 
where $d$ is the Euclidean distance between the superpixel features $F_{i'}^A$ and
$F_{j'}^B$, 
for instance, average superpixel color or 
normalized cumulative color histogram,
and $w$ is a weight depending on the relative position of
the superpixels $A_{i'}$ and $B_{j'}$. 
Note that we consider an Euclidean distance, but any distance on superpixel features can be computed with $d$.\smallskip

\subsubsection{Fast distance between superpixels}
To compute the weight between two superpixels $A_{i'}$ and $B_{j'}$,
we would ideally like to measure their
relative overlapping area, 
\emph{i.e.}, setting
$w(A_{i'},B_{j'})=|A_{i'}\cap B_{j'}|/|A_{i'}\cup B_{j'}|$. 
Nevertheless, this computation requires the expensive count of overlapping
pixels that cancels the computational advantage of the superpixel representation.
A fast method would be to compare 
a superpixel $A_{i'}$ to the
spatially closest $B_{j'}$, but
we propose a more robust framework that
considers
a spatial distance between the superpixel barycenters.
We define the symmetric spatial weight between two superpixels $A_{i'}$ and $B_{j'}$ as: 
{\eqsize
\begin{equation}
\label{weight}
w(A_{i'},B_{j'}) =\exp\left(-x_{i'j'}^Tx_{i'j'}/\sigma_1^2\right)w_s(A_{i'})w_s(B_{j'}) ,
\end{equation}
}\noindent where $x_{i'j'}=c_{j'}-c_{i'}+v_{ij}$ is the relative distance between
$c_{i'}$ and $c_{j'}$,
$w_s(A_{i'})$ weights the influence of $A_{i'}$ according to 
its spatial distance to $A_i$ such that $w_s(A_{i'})=\exp(-\|c_i - c_{i'}\|_2^2/\sigma_2^2)$, and
$\sigma_1$ and $\sigma_2$ are two scaling parameters.
The setting of 
$\sigma_1$ depends of the superpixel decomposition scale.
Since the superpatches have been registered, and the
aim is to compare a superpixel $A_i\in \spatch{A_i}$ to the closest ones $B_j\in \spatch{B_j}$,
$\sigma_1$ can be set to half the average superpixel size, \emph{i.e.}, half the average distance between superpixel barycenters, 
such that $\sigma_1=\frac{1}{2}\sqrt{h{\times}w/K}$, 
for an image of size $h{\times}w$ pixels decomposed into $K$ superpixels.
Finally, $\sigma_2$ depends on the superpatch size and can be set to $\sqrt{2}R$
to weight the contribution of closest superpixels.
\smallskip

\subsubsection{Generalization of pixel-based patches}
In the limit case where each superpixel
only contains one pixel, \emph{i.e.}, $A_{i'}$=$c_{i'}$, $B_{j'}$=$c_{j'}$, 
$\spatch{A_i}$ and $\spatch{B_j}$ have the same regular structure and the same number of elements. 
With $\sigma_1 \to 0$ and $\sigma_2 \to \infty$ in \eqref{weight},
$w(A_{i'},B_{j'})=w(c_{i'},c_{j'})=1$ if $c_{i'}-c_i=c_{j'}-c_j$ and $0$ otherwise, and
the proposed
distance $D$ \eqref{D} is a generalization of the distance between patches, since it 
reduces to a normalized standard SSD between two pixel-based patches: 
{\eqsize
\begin{align}
  D(\spatch{A_i},\spatch{B_j}) &= \frac{1}{|\spatch{A_i}|}\sum_{i'\in \mathcal{I}^A_i,j'\in \mathcal{I}^B_j}^{}  d(F_{i'},F_{j'}) \delta_{c_{i'}-c_i=c_{j'}-c_j}   , \nonumber\\ 
  &= \text{SSD}(\spatch{A_i},\spatch{B_j})  , 
  \end{align}
}\noindent where
$\delta_{a}=1$ when $a$ is true and $0$ otherwise.

 \section{\label{section:superpatchmatch}SuperPatchMatch}
 \subsection{\label{subsection:superpatchmatch}The SuperPatchMatch Algorithm}

We propose the SuperPatchMatch method (SPM),
an extension of the PatchMatch (PM) algorithm \cite{barnes2009} dedicated
to our superpatch framework,
for fast matching of irregular structures
from superpixel decompositions.
In this section, only direct adjacent neighborhood relationship needs to be considered 
to design our algorithm.
Nevertheless, as for pixels described by regular patches, 
we demonstrate in Section \ref{section:expe} that the proposed framework is
significantly more efficient using superpatches.
In the following, as in Section \ref{subsection:SuperPatchComparison}, 
we illustrate the proposed method by considering two superpatches
$\spatch{A_i}$ and $\spatch{B_j}$, in different images $A$ and $B$,
but our approach can be applied to an
entire database, as demonstrated in Section~\ref{section:expe}.

PM is a method that computes pixel-based patch
correspondences between two images. 
The key point of this method is that 
good correspondences can be propagated to the adjacent
patches within an image.
The algorithm has three steps: initialization, propagation
and random search.  
The initialization consists in randomly associating each
patch of the image $A$ with a patch of the image $B$,
leading to an initial ANN field.   
The following steps are then iteratively performed to improve
the correspondences.
The propagation uses
the assumption that when a patch in $A$ corresponds to a
patch in $B$, then the adjacent patches in $A$ should
also match  the adjacent patches in $B$.
The random search consists in
a sampling around the current ANN to escape from possible
local minima.  

The lack of regular geometry between superpatches is the main
issue for adapting
the PM algorithm. The notion of adjacent patches has to be defined for an
irregular superpixel decomposition.  
For the sake of clarity, the current best ANN of a superpixel $A_i\in A$,
is denoted as $B(i)=B_{\mathcal{M}(i)}$,
with $\mathcal{M}$ the ANN map which stores, for superpixels in $A$, 
the index in $B$ of their corresponding ANN.\smallskip

\subsubsection{SPM initialization step}
For each superpixel $A_i\in A$, we assign a random superpixel $B(i)\in B$.
Fig. \ref{fig:SPPMa} shows initialization examples.
After this step, 
propagation and random search are
iteratively performed to improve the initial
matches.

\newcommand{\sizpm}{72pt}

\begin{figure}[ht]
\centering
\newcommand{\siz}{0.95\columnwidth}
\includegraphics[width=\siz,height=\sizpm]{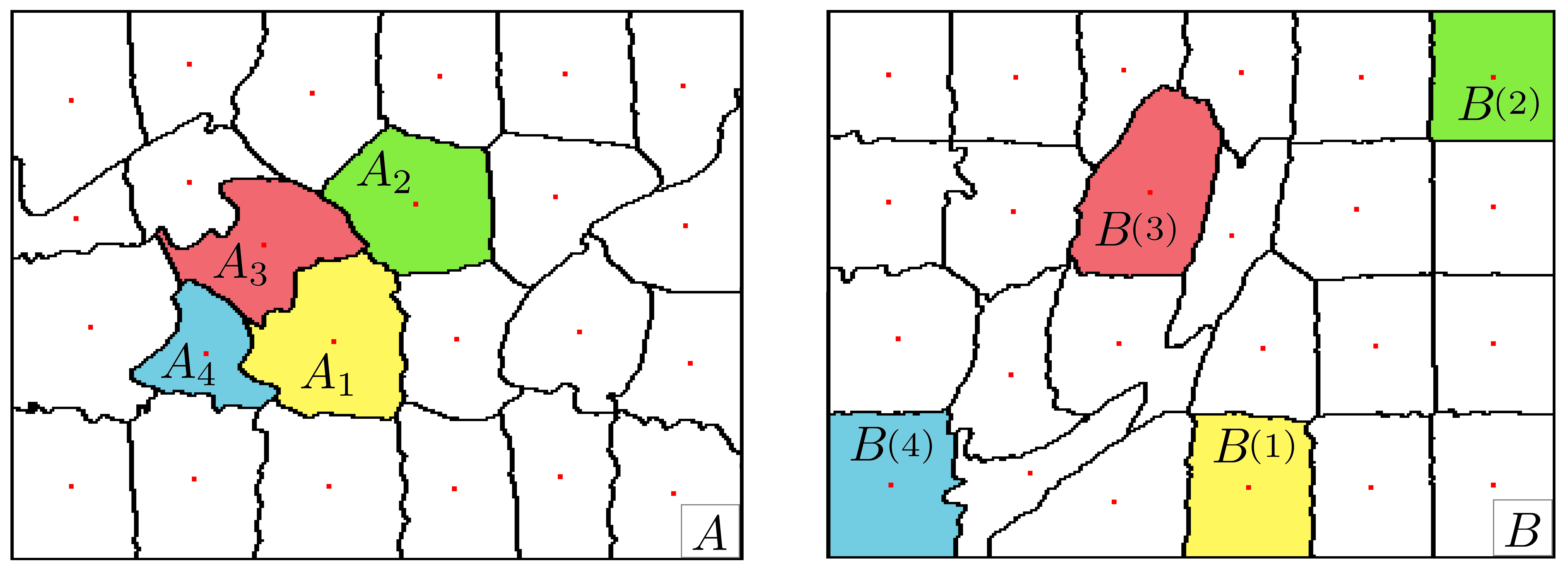}
\caption{SPM initialization step. Each superpixel $A_i\in A$
  is randomly assigned to a superpixel $B(i)\in B$.}
\label{fig:SPPMa}
\end{figure}

\subsubsection{SPM propagation step}
\begin{figure}[ht]
\centering
\newcommand{\siz}{0.95\columnwidth}
\includegraphics[width=\siz,height=\sizpm]{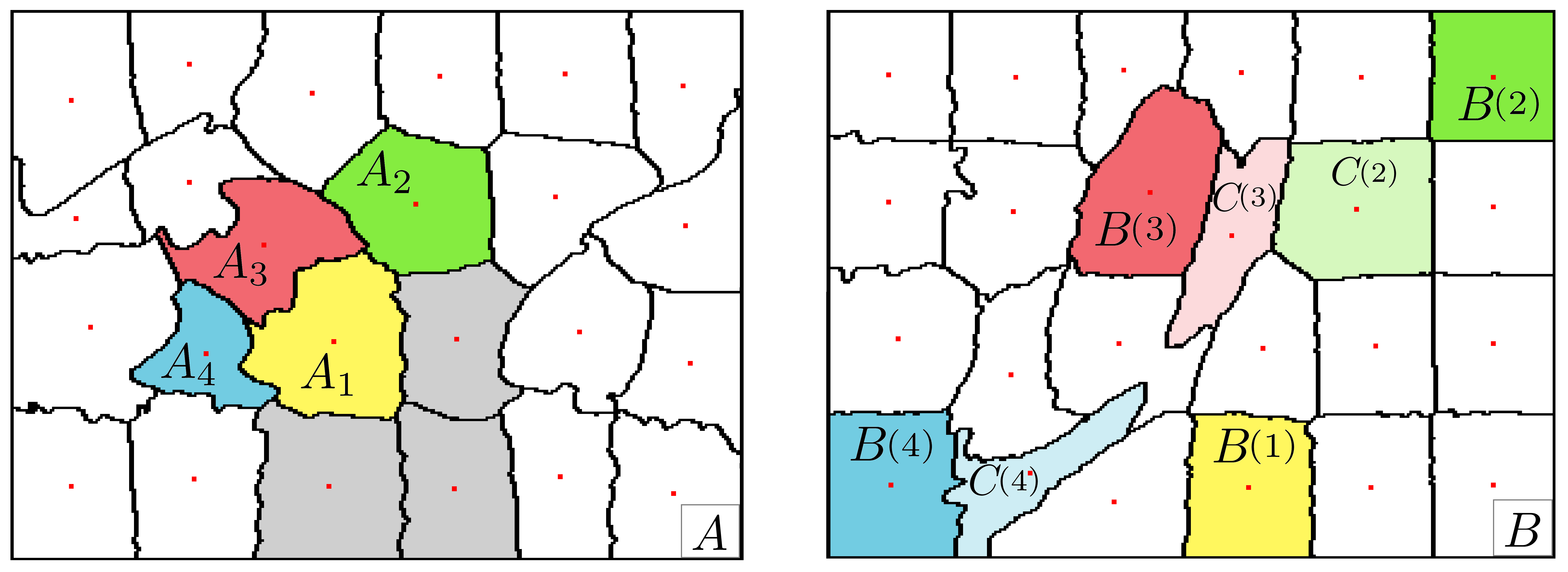}
\caption{SPM propagation step. 
 According to the
 scan order,
only top-left superpixels of $A_1$ are considered on even iterations
 ($A_2$, $A_3$ and $A_4$). 
The remaining neighbors (in gray) are tested on odd iterations.
Current matches are denoted by $B(i)$, while $C(i)$ represent the new candidates to test as ANN.}
\label{fig:SPPMb}
\end{figure}

In \cite{barnes2009}, the propagation step tries to improve 
the current ANN by using the ones of the directly adjacent neighbors in $A$.
Pixels are processed according to a scan order,
\emph{e.g.}, from top-left to bottom-right.
The propagation only considers previously processed 
and directly adjacent neighbors, \emph{e.g.}, top and left.
Their ANN are shifted to respect the relative position of pixels in $A$, 
providing ANN candidates to the currently processed pixel.
With superpixels, the selection of top, left, bottom
and right adjacent superpixels is not direct,  
since there is no regular geometry between them. 
We propose to define the superpixel scan order from the raw pixel order on $A$ (left to right, top to bottom),
and to consider in the propagation step
all adjacent superpixels that were processed during the current iteration.
The selection of candidates from adjacent neighbors is illustrated in
Fig. \ref{fig:SPPMb}.

\begin{figure}[t!]
\centering
\newcommand{\siz}{0.95\columnwidth}
\includegraphics[height=105pt, width=\siz]{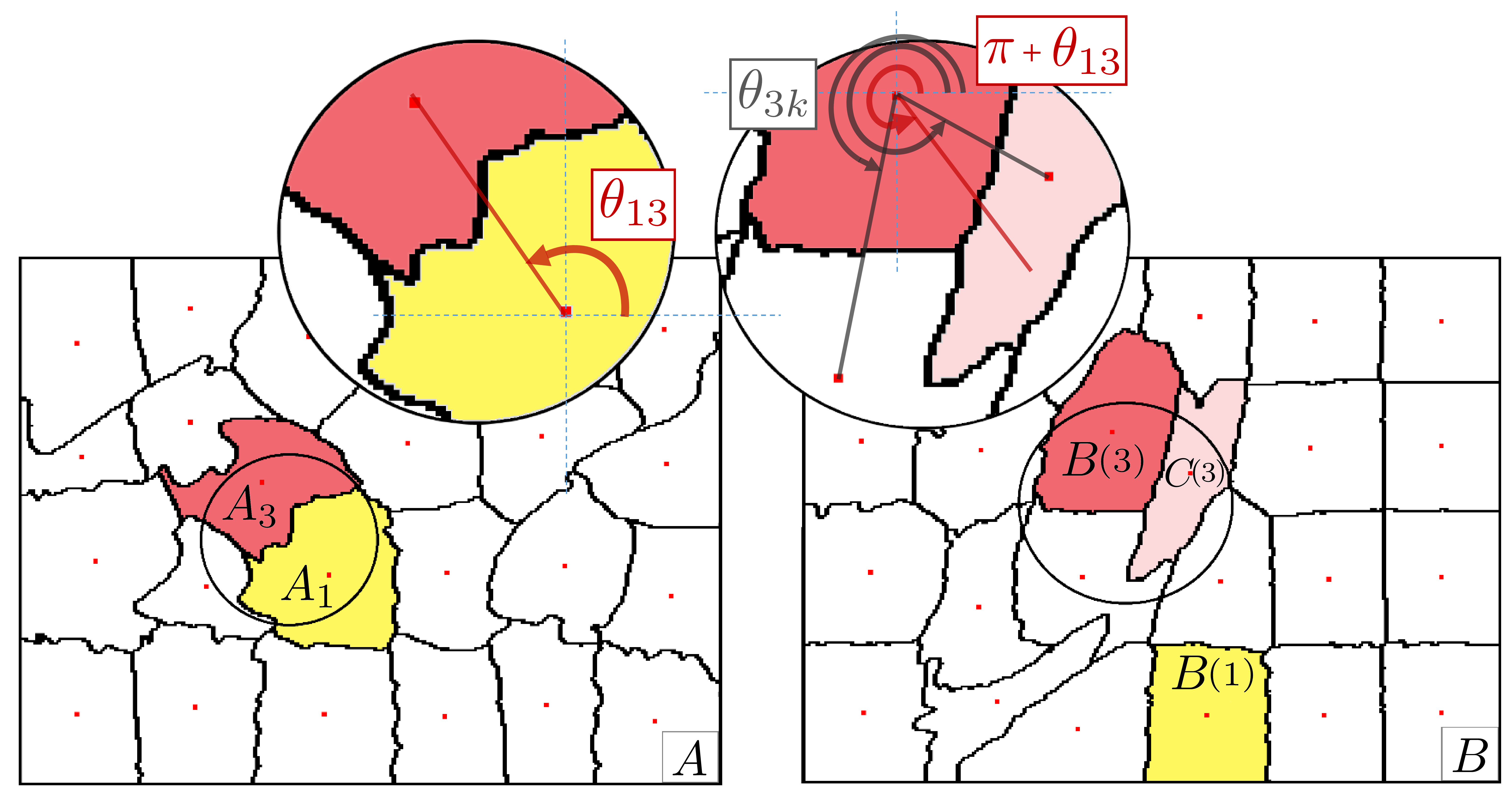}
\caption{
To improve the ANN of the superpixel $A_1$, the neighbor
$A_{3}$ is considered.  
Its current ANN is $B(3)$. 
The selected superpixel $C(3)$ to test is the adjacent superpixel of $B(3)$ with the most similar
symmetric orientation, \emph{i.e.}, 
the superpixel $B_k$ which angle to $B(3)$,
$\theta _{3k}$, 
is the closest to $\pi + \theta_{13}$.}
\label{fig:prop_detail}
\end{figure}

When an adjacent superpixel $A_{i'}$ and its ANN $B(i')$ are considered, 
a neighbor of $B(i')$ is selected as a candidate to improve the correspondence of $A_i$.
However, the ANN $B(i')$ cannot be shifted as done for regular patches, 
since the superpixels are defined on irregular domains. 
Therefore, to improve the ANN of $A_i$, 
one particular neighbor of $B(i')$, denoted as $C(i')\in B$ is tested. 
It is given by the superpixel whose relative position to $B(i')$ 
is the most similar to $\theta_{ii'}$, the angle between $A_i$ and $A_{i'}$.
Hence, the ANN candidate $C(i')$ to test is obtained as: 
{\eqsize
\begin{equation}
C(i')=\underset{k\in \mathcal{N}^B_{\mathcal{M}(i')}}{\argmin}  |(\thetaiiprim+\pi ) - \theta _{i'k} |,
\end{equation}
}\noindent with $\theta _{i'k}$ the angle between $B(i')$ and
its set $\mathcal{N}^B_{\mathcal{M}(i')}$ of adjacent superpixels $B_k$. 
Note that all angles are computed from superpixel barycenters.
The selection of the candidate  for $A_{3}$, which is on top-left position of $A_1$,  is illustrated in Fig. \ref{fig:prop_detail}. 
To keep the same relative position, the new superpixel to test as ANN for $A_1$ is the neighbor of $B(3)$ that is the closest to its bottom-right position,
according to the angle $\theta _{i'k}$. \smallskip

\subsubsection{SPM random search step}
\begin{figure}[ht]
\centering
\includegraphics[width=0.95\columnwidth,height=\sizpm]{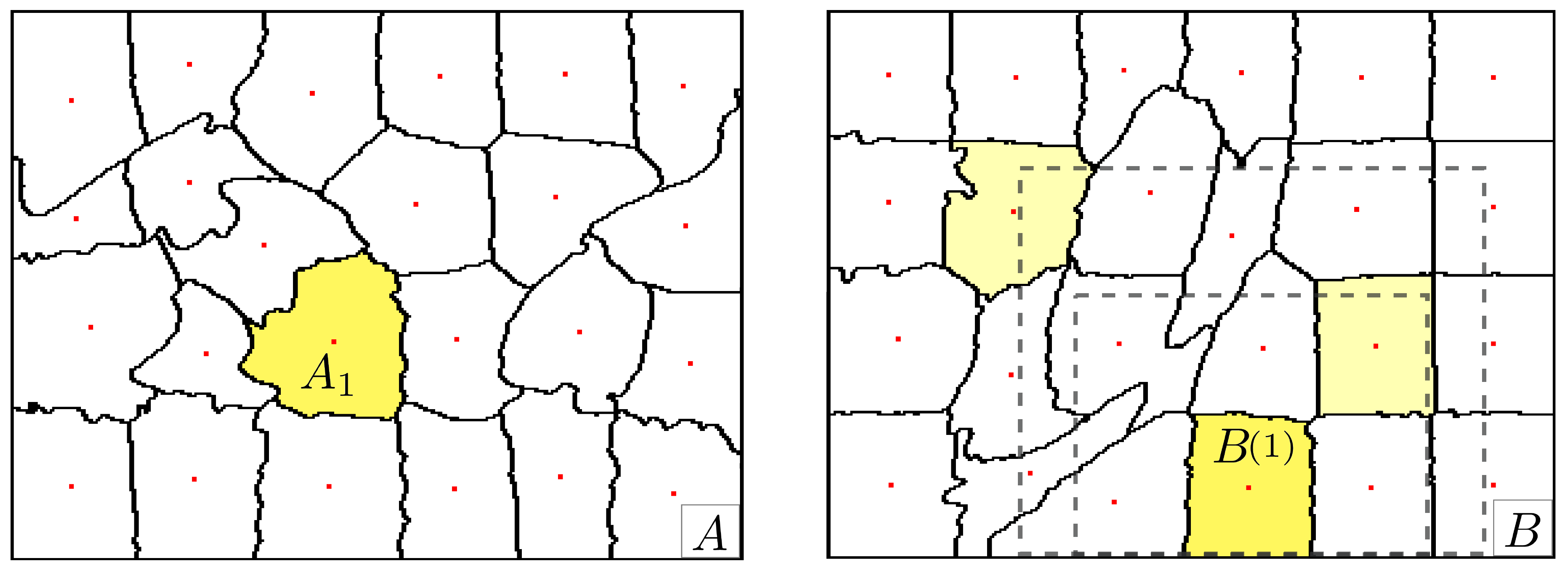}
\caption{SPM random search step.
The sampling is performed at a decreasing distance around the barycenter of the ANN $B(1)$
of $A_1$. 
The superpixels containing the selected positions (crosses) are the candidates to test.
}
\label{fig:SPPMc}
\end{figure}

The random search step consists in a sampling around the current ANN 
to escape from possible local minima \cite{barnes2009}. 
Candidates are selected at an
exponentially decreasing  
distance from the barycenter of the best current match.
A random pixel position is computed within decaying boxes, and
the superpixel containing this pixel is the candidate to test.
Fig. \ref{fig:SPPMc} illustrates the random search step, where the boxes are depicted in dotted lines.

\subsection{Library of Training Images}
One advantage of the SPM algorithm is that its complexity
only depends on the size of the image to process and not on
the  size of the compared image database.
This important fact enables SPM to perform fast ANN searches within 
a large database with no increase on the computational time.

All example images within the database are grouped into a single library $T$. 
In the case of a training database, SPM steps are adapted so the ANN can be found within all images.
The initialization is extended: the ANN is randomly selected within $T$.
The propagation step still tests the shifted ANN of the neighbors, that are not necessarily in the same training image.
Finally, the random search is performed
within the current best image, and 
within a random image in $T$, as in
\cite{shi2013}.

\subsection{Multiple SPM}
Contrary to PM, that only estimates one ANN,
SPM computes $k$-ANN matches in the library $T$, since
the diversity of information from multiple ANN may
help to perform more accurate processing.
In the literature, an extension of the original PM algorithm
to the $k$-ANN 
case has been proposed in \cite{barnes2010}. The suggested
strategy is to build a constantly updated data structure of the best
visited correspondences.  
However, to parallelize such an approach, the current test image must
be split into several parts which leads to boundary issues.
Therefore, we chose to implement the $k$-ANN
search by $k$ fully
independent SPM, leading to a simpler scheme.

\section{Application to Image Labeling\label{section:expe}}

To demonstrate the interest
of the superpatch structure and the
SPM algorithm, we adapt our
approach to
exemplar-based
labeling.
We consider two experiments: 
face labeling on the LFW dataset \cite{huang2007db},
and segmentation and labeling on non-registered medical images from the BRATS dataset \cite{brats2012}.

\subsection{Label Fusion Method}

The proposed algorithm is 
particularly interesting for labeling applications.
The superpixel decomposition segments the image into
homogeneous regions that try to respect existing contours, 
and SPM finds superpixel-based correspondences whose labels can be transfered.
In this application, a library $T$ of training images with their label ground truths is considered, 
and SPM provides $k$-ANN matches.
We denote as $l(T_j)$ $\in$ $\{1,\dots ,M\}$ the label of the training superpixel $T_j$ contained in $T$.
The labels of the selected ANN within $T$ are merged by a patch-based label fusion \cite{coupe2011patch}, inspired from \cite{buades2005}.

At the end of the ANN search, $k$-ANN are estimated for all
superpixels in the test image $A$. 
To obtain the final labeling, 
for a superpixel $A_i$ 
and $\KK _i^m= \{T_j\}$ the set of its $k$-ANN matches with label $l(T_j)=m$, 
its label fusion map $L_m(A_i)$ is defined by: 
{\eqsize
\begin{equation}
L_m(A_i) =\frac{\sum_{T_j\in \KK _i^m} \omega(A_i,T_j)} 
            {\sum_{m=1}^M\sum_{T_j\in \KK _i^m} \omega(A_i,T_j)}\, ,
\label{labelfusion}
\end{equation}
}\noindent where $\omega(A_i,T_j)$ is the weight contributing to label $m$, and
depends on the similarity between the superpatch $\spatch{A_i} \in A$, 
and the ANN superpatch $\spatch{T_j} \in T$.
This label map $L_m$ gives the probability of assigning the label $m$ to the superpixel $A_i$. 

Some applications can also deal with registered images, 
where structures of interest between $A$ and images of $T$ are spatially close. 
Therefore, good superpatch matches should not be spatially too far in the image domain.  
In this case,
to enforce the spatial coherency of the selected $k$-ANN, 
each labeling contribution is weighted by 
the spatial distance between the central superpixels barycenters $c_i\in A$ and $c_j\in
T$: 
{\eqsize
\begin{equation}
\omega(A_i,T_j) =
\exp{\left(1-\left(\frac{D(\spatch{A_i},\spatch{T_j})}{h(A_i)^2} 
    + \frac{\|c_i - c_j\|_2}{\beta ^2}\right)\right)} , 
    \label{finalweight}
\end{equation}
}\noindent where
{\small
$h(A_i)^2=\alpha ^2 \underset{T_j\in \underset{{\sss m}}{\cup}\KK _i^m}{\min}(D(\spatch{A_i}
,\spatch{T_j} ) + \epsilon ) $, 
}
with $\epsilon\to 0$, and
$\alpha $ and $\beta$ are scaling parameters. 
With the function $h(A_i)$, the distance of the current contribution
is divided by the minimal distance among all $k$-ANN contributions.
 For each superpixel $A_i$, the final labeling map $\LL(A_i)$ is obtained with the label of highest probability:
{\eqsize
\begin{equation}
  \LL (A_i ) = \underset{m\in {\{1,\dots ,M\}}}{\argmax}L_m(A_i).
  \label{finaldecmulti} 
\end{equation}
}
The relation \eqref{finaldecmulti} gives a superpixel-wise decision that may have some irregularities.
As in \cite{pei2014}, we can use
\eqref{labelfusion} as a multi-label data term and consider the following regularization problem, 
that consists in minimizing the energy $J$, defined on the graph built from adjacent superpixels: 
{\eqsize
\begin{align}
J(\LL)\hspace{-0.05cm}=\hspace{-0.05cm}\sum_{i=1}^{|A|} \hspace{-0.05cm}\left( \sum_{i'\in \mathcal{I}^A_i}\hspace{-0.1cm}\exp{\left(\hspace{-0.05cm}-\frac{d(F_i,F_{i'})}{\gamma}\hspace{-0.05cm}\right)} \delta_{i,i'}
     \hspace{-0.05cm}+\hspace{-0.05cm}1\hspace{-0.05cm}-\hspace{-0.05cm}L_{\LL(A_i)}(A_i)\hspace{-0.1cm}\right), \label{gc}
\end{align}
}\noindent where $\gamma$ is a regularization parameter, the data term 
$1-L_{\LL(A_i)}(A_i)$ is close to $0$ (respectively $1$) when the probability of label $\LL(A_i)$ is high (respectively low),
and $\delta_{i,i'}=1$ when $\LL(A_i) \neq\LL(A_{i'})$ and $0$ otherwise. 

\begin{figure*}[t]
\centering
\newcommand{\sizp}{0.19\textwidth}
\newcommand{\sizpp}{0.15\textwidth}
{\footnotesize
\begin{tabular}{@{\hspace{0mm}}c@{\hspace{1mm}}c@{\hspace{7mm}}c@{\hspace{0mm}}c@{\hspace{0mm}}c@{\hspace{0mm}}}
\includegraphics[width=\sizp,height=\sizpp]{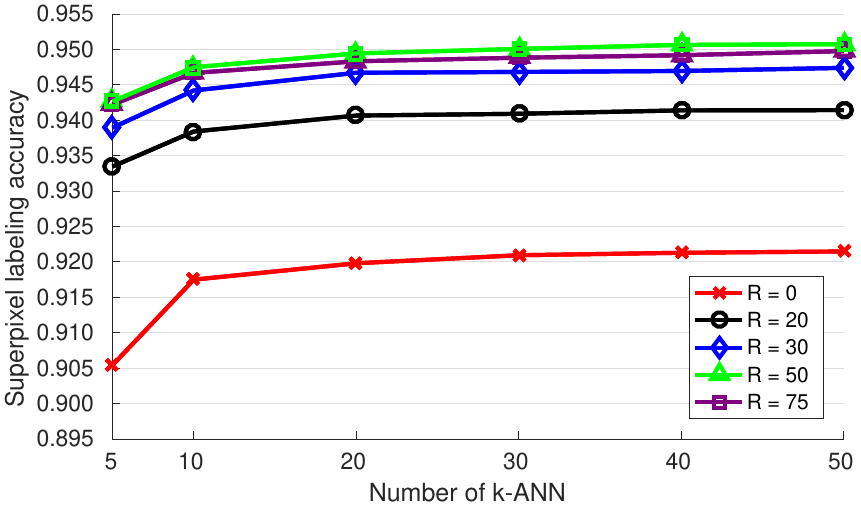} &
\includegraphics[width=\sizp,height=\sizpp]{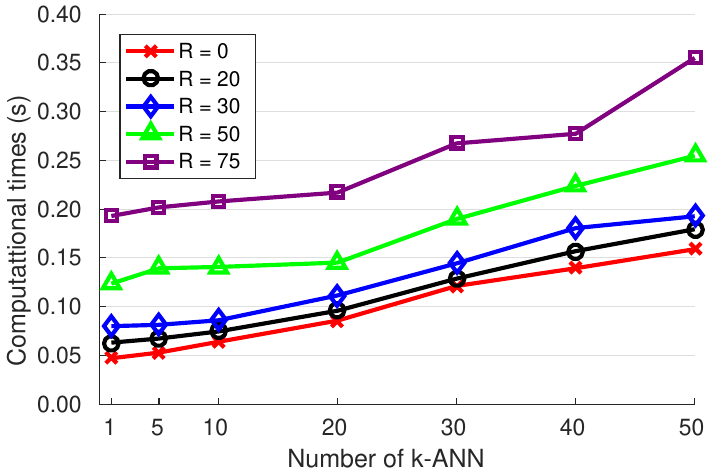}& 
\includegraphics[width=\sizp,height=\sizpp]{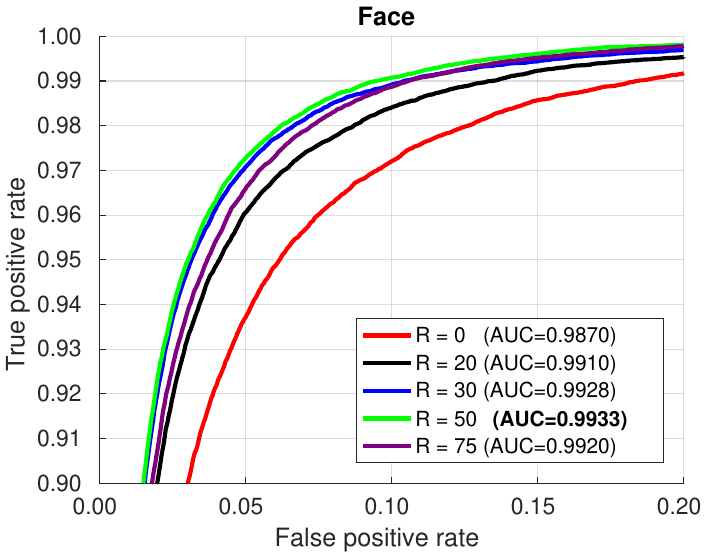} &
\includegraphics[width=\sizp,height=\sizpp]{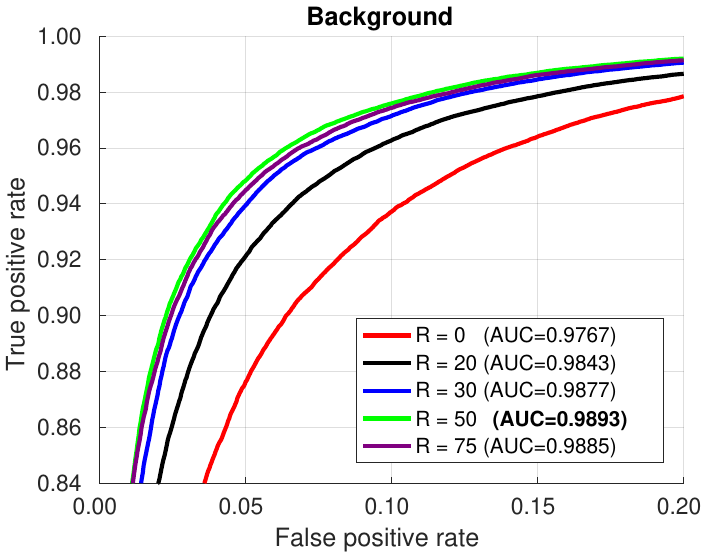}&
\includegraphics[width=\sizp,height=\sizpp]{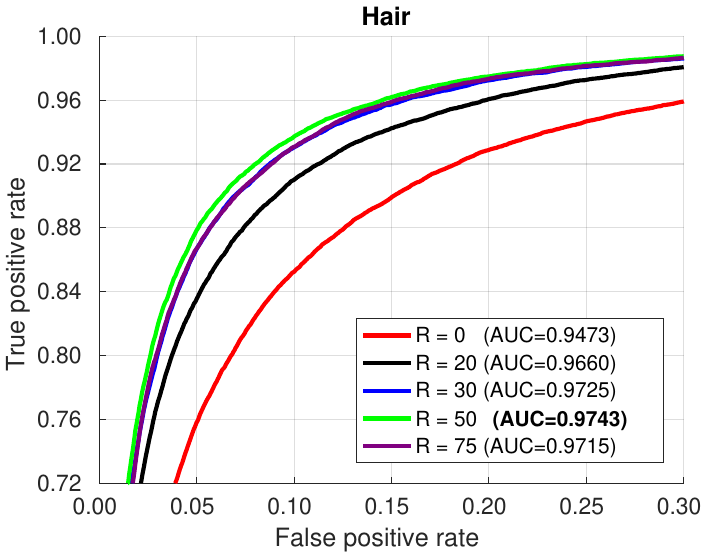}\\
(a)&(b)&(c)&(d)&(e)\\
\end{tabular}
}
\caption{Influence of the superpatch size on
superpixel-wise labeling accuracy (a) with corresponding 
computational time (b).
ROC curves with area under curves (AUC) for several superpatch sizes ($k$=$50$)
for face (c), background (d) and hair (e).
$R=50$ pixels gives the best results for all labels.
}
\label{fig:roc_curves}
\end{figure*}

\subsection{Face Labeling Experiments\label{section:xps}}

Face segmentation and labeling are challenging tasks due to several
issues such as 
the diversity of hair styles, background, color skins, or occlusions.
We evaluate the proposed SPM approach for face
labeling on the funneled version of the Labeled Faces in the Wild (LFW)
dataset \cite{huang2007db}.
The dataset contains 2927 images of size $250{\times}250$ pixels, 
that have been coarsely aligned
\cite{huang2007}, and
segmented into
225 to 250 superpixels.
LFW is a widely used database for validating new methods based on superpixels
since it contains decompositions with associated superpixel-wise ground truths,
and comparisons with state-of-the-art 
methods are not biased by the ground truth superpixel decomposition one would have to compute. \smallskip

\subsubsection{Parameter settings}
SPM was implemented with MATLAB using
C-MEX code.
Our experiments are performed on a standard Linux server of 16 cores
at 2.6 GHz with 100 GB of RAM.  
To compare to \cite{kae2013,liu2015}, 
we use the same 1 500 training images, and the same 927 images for testing.
Nevertheless, we could use all images in a 
leave-one-out procedure since our method does not need any training step.

The number of SPM iterations is set to 5, as in \cite{barnes2009}.
We only use a $\ell_2$-norm between histogram of oriented gradients (HoG) \cite{dalal2005} as distance $d$ in \eqref{D}.
In Eq.~\eqref{weight}, 
since the images are $h{\times}w = 250{\times}250$ pixels, and decomposed into approximately $K=250$ superpixels, 
$\sigma_1$ is set to $8$.
In Eq.~\eqref{finalweight}, parameters $\alpha$ and $\beta$ are
respectively set to 2 and 4. 
Finally, we set $\gamma$ to 0.5 and
use the $\alpha$-expansion algorithm \cite{boykov2001} to minimize  \eqref{gc}.
The reported times for SPM in Fig. \ref{fig:roc_curves} (b)
include $k$-ANN searches, label fusion 
and the complete labeling with regularization. \smallskip

\subsubsection{Influence of the superpatch size}
We first investigate the influence of the superpatch size and number
of ANN.
Fig. \ref{fig:roc_curves} represents the superpixel-wise labeling
accuracy and computational time.
The labeling accuracy is increased with our superpatch structure.
Best results are obtained with $R$=$50$ pixels ($95.08\%$ with $k$=$50$ ANN).
Such superpatch size corresponds in average to the capture of the three neighboring rings of superpixels, 
since superpixels are approximately of size $16{\times}16$ pixels.
Fig. \ref{fig:roc_curves} also represents the corresponding ROC curves obtained with $k$=$50$ ANN
for the three classes (face, background, hair).
Without the superpatch structure, \emph{i.e.},
only computing the distance
on central superpixels ($R$=$0$ pixels),
worse ANN are found, decreasing the labeling accuracy ($93.29\%$ with $k$=$50$ ANN).
The superpatch size must be large enough to capture the information contained within the superpixel
neighborhood. 
However, with too large superpatches, \emph{i.e.}, $R$ $>$ $50$ pixels, too many neighboring superpixels
contribute, leading
to less relevant ANN and less accurate labeling. 
Note that we propose in \eqref{finalweight} a slight improvement of the label fusion step to take into account the
LFW database registration but we obtain very comparable results ($95.00\%$ instead of $95.08\%$)
without any position a priori, \emph{i.e.}, $\beta$=$\infty$ in \eqref{finalweight}.

Fig. \ref{fig:proba} illustrates the regularization process.
Labeling probabilities \eqref{labelfusion} obtained from SPM are displayed for each label (Fig. \ref{fig:proba}(d)).
The spatial regularization \eqref{gc} gives more consistent results (Fig. \ref{fig:proba}(f)) than taking the label of highest probability (Fig. \ref{fig:proba}(e)). 
Finally, Fig. \ref{fig:LFW}  shows the superpatch influence
on labeling for various examples.
Labeling failures are mostly due to high similarity between hair and background, 
or inaccurate superpixel segmentation. \smallskip

\newcommand\siz{0.115\textwidth}
\newcommand\sizz{0.115\textwidth}
\begin{figure*}[ht!]
\begin{scriptsize}
\centering 
\begin{tabular}{@{}c@{}c@{}c@{\hspace{3mm}}c@{}c@{}c@{\hspace{3mm}}c@{}c@{}}   
 && & & & &$94.40\%$&$97.60\%$\\
\includegraphics[width=\siz,height=\sizz]{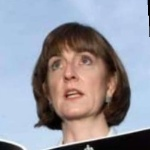}&
\includegraphics[width=\siz,height=\sizz]{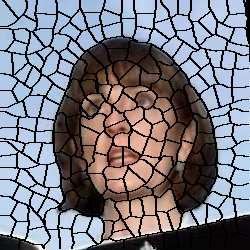}&\includegraphics[width=\siz,height=\sizz]{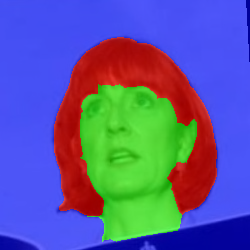}&\includegraphics[width=\siz,height=\sizz]{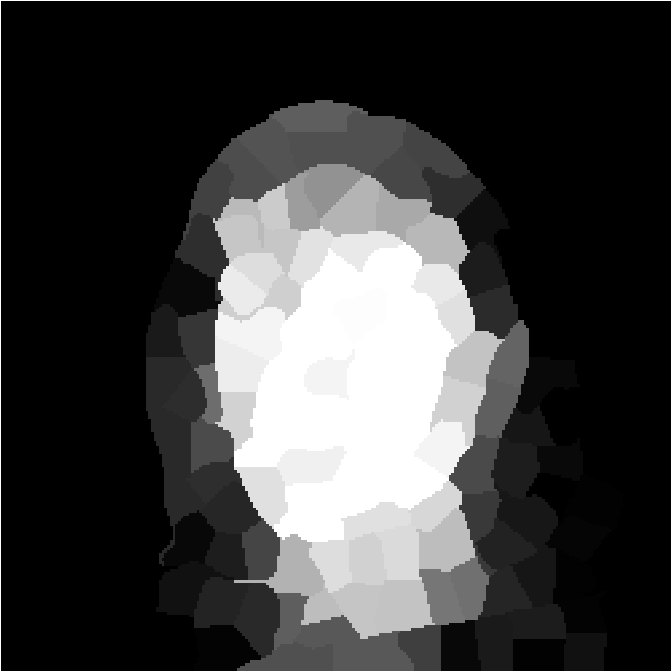}&\includegraphics[width=\siz,height=\sizz]{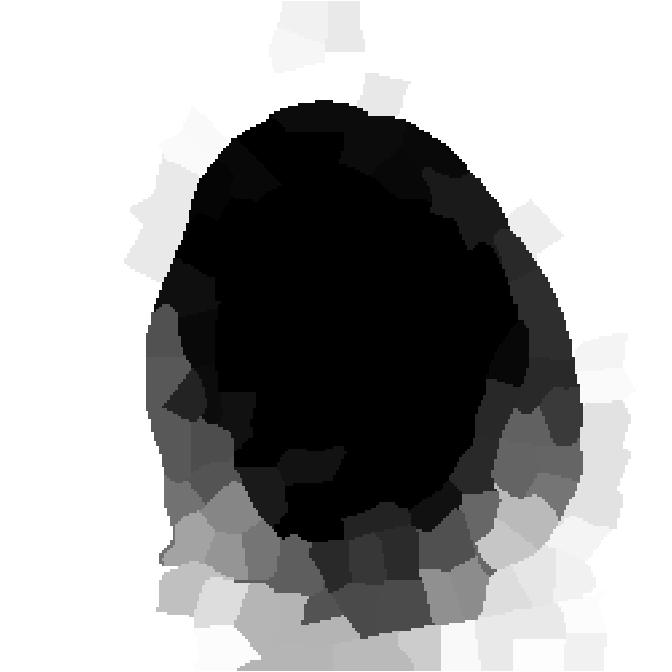}
&\includegraphics[width=\siz,height=\sizz]{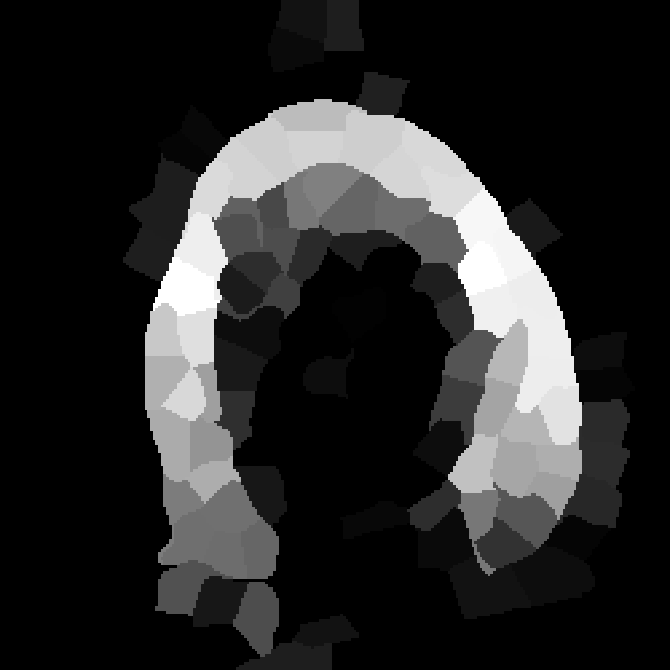}&\includegraphics[width=\siz,height=\sizz]{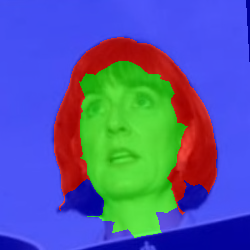}&\includegraphics[width=\siz,height=\sizz]{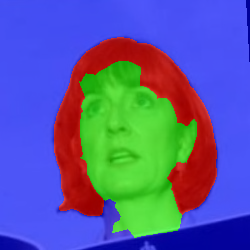}\\
 && & & & &$91.23\%$&$96.41\%$\\
 \includegraphics[width=\siz,height=\sizz]{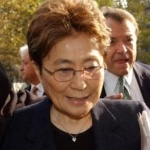}&
\includegraphics[width=\siz,height=\sizz]{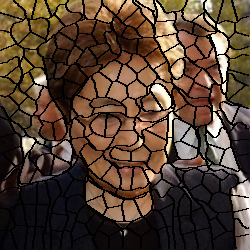}&\includegraphics[width=\siz,height=\sizz]{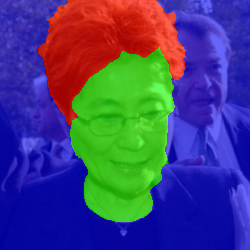}&\includegraphics[width=\siz,height=\sizz]{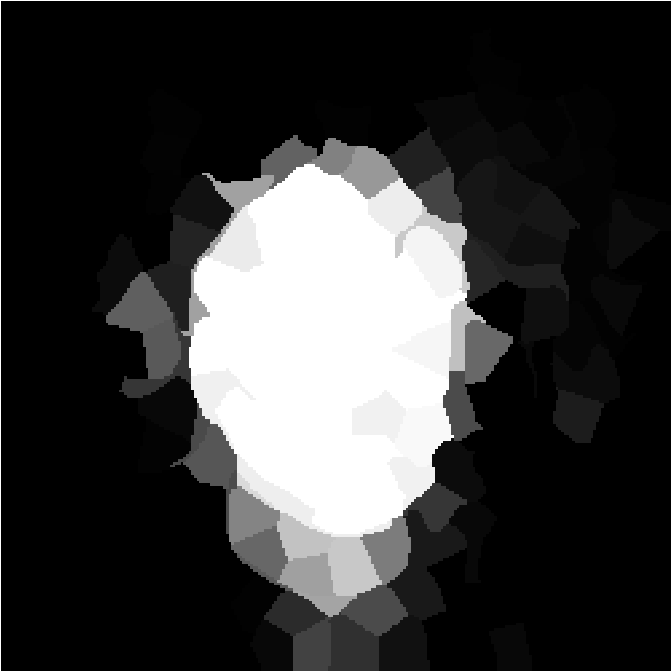}&\includegraphics[width=\siz,height=\sizz]{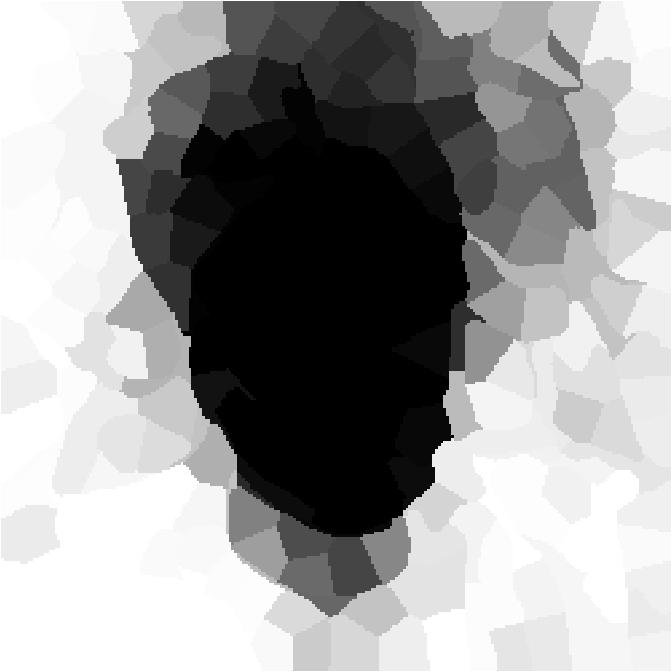}
&\includegraphics[width=\siz,height=\sizz]{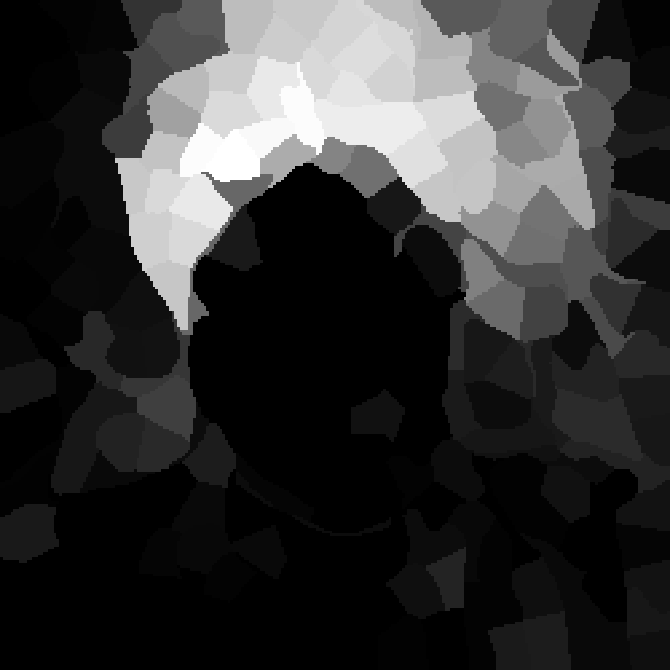}&\includegraphics[width=\siz,height=\sizz]{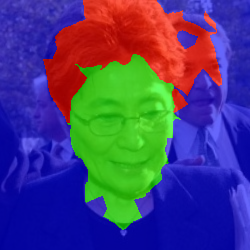}&\includegraphics[width=\siz,height=\sizz]{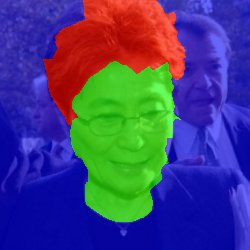}\\
(a)&(b)&(c)&\multicolumn{3}{c}{(d) Labeling probabilities \hspace{0.5cm}}&(e)&(f)\\
Image&Superpixels&Ground truth&Face&Background&Hair&Highest probability&Regularization \\
\end{tabular}
\caption{
Labeling results with SPM. (a) Image. (b) Superpixel decomposition. (c) Associated ground truth. (d) Label fusion maps for the 3 classes (face, background, hair). 
(e) Labeling with the highest probability from \eqref{finaldecmulti}. (f) Labeling after regularization.
}
\label{fig:proba}
\end{scriptsize}
\end{figure*}

\newcommand\zisz{0.078\textwidth}
\begin{figure*}
\begin{scriptsize}
\centering 
\begin{tabular}{@{\hspace{1mm}}c@{}c@{}c@{}c@{\hspace{1mm}}} 
&&$60.47\%$&$96.90\%$\\
\includegraphics[width=\zisz]{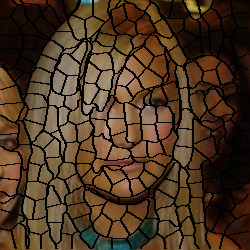}&\includegraphics[width=\zisz]{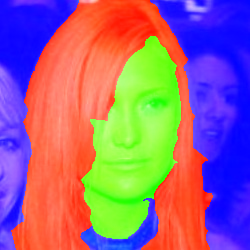}&
\includegraphics[width=\zisz]{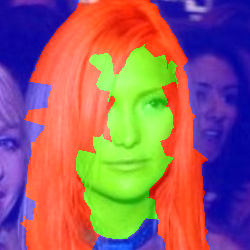}&
\includegraphics[width=\zisz]{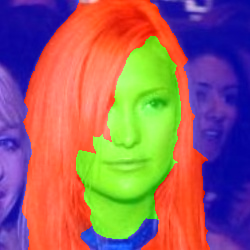}\\
&&$89.35\%$ 
&$96.96\%$\\
\includegraphics[width=\zisz]{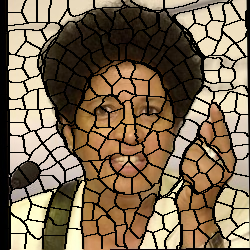}&\includegraphics[width=\zisz]{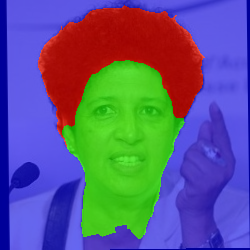}&
\includegraphics[width=\zisz]{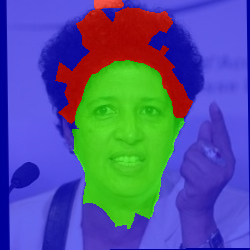}
&\includegraphics[width=\zisz]{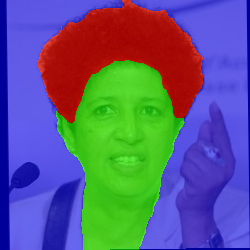}\\
&&$95.24\%$ 
&$98.81\%$\\
\includegraphics[width=\zisz]{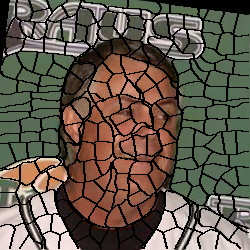}
&\includegraphics[width=\zisz]{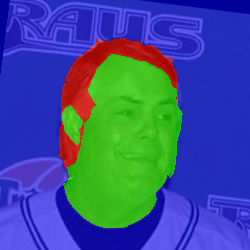}&
\includegraphics[width=\zisz]{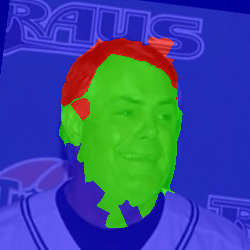}
&\includegraphics[width=\zisz]{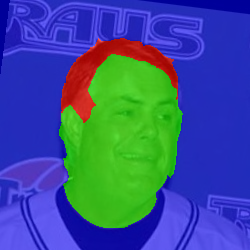}\\
Superpixels&Ground truth&SPM $R$=$0$&SPM $R$=$50$
\end{tabular}
\begin{tabular}{@{\hspace{1mm}}c@{}c@{}c@{}c@{\hspace{1mm}}} 
&&$96.06\%$
&$99.61\%$\\
\includegraphics[width=\zisz]{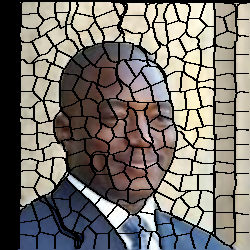}&
\includegraphics[width=\zisz]{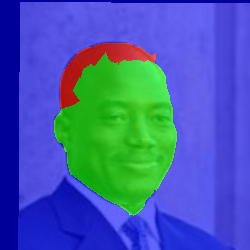}&
\includegraphics[width=\zisz]{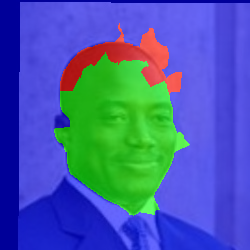}&
\includegraphics[width=\zisz]{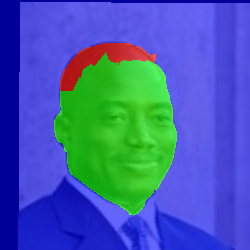}\\
&&$93.46\%$ 
&$97.69\%$\\
\includegraphics[width=\zisz]{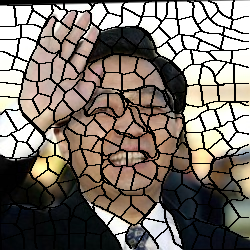}&
\includegraphics[width=\zisz]{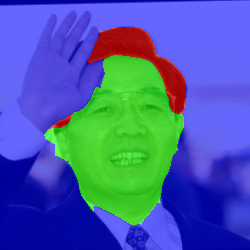}&
\includegraphics[width=\zisz]{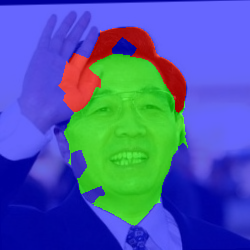}
&\includegraphics[width=\zisz]{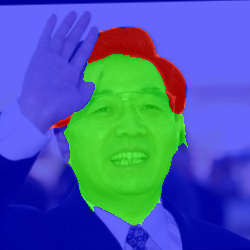}\\
&&$80.53\%$ 
&$97.33\%$\\
\includegraphics[width=\zisz]{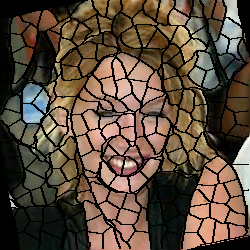}
&\includegraphics[width=\zisz]{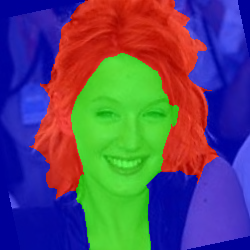}&
\includegraphics[width=\zisz]{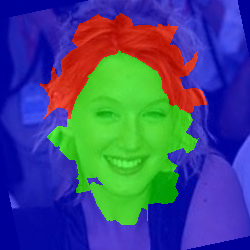}
&\includegraphics[width=\zisz]{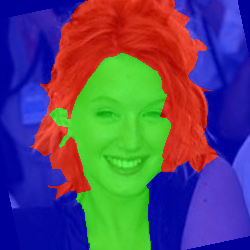}\\
Superpixels&Ground truth&SPM $R$=$0$&SPM $R$=$50$
\end{tabular}
\begin{tabular}{@{\hspace{1mm}}c@{}c@{}c@{}c@{\hspace{1mm}}} 
&&$75.69\%$ 
&$93.33\%$\\
\includegraphics[width=\zisz]{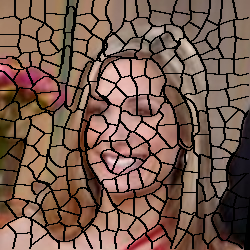}&
\includegraphics[width=\zisz]{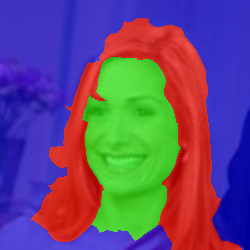}&
\includegraphics[width=\zisz]{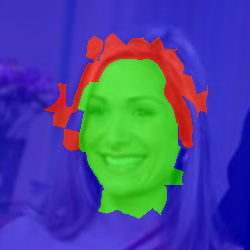}
&\includegraphics[width=\zisz]{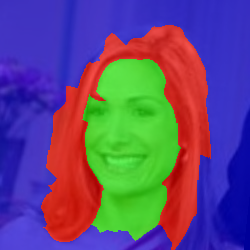}\\
&&$94.16\%$ 
&$98.05\%$\\
\includegraphics[width=\zisz]{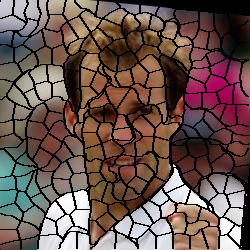}&
\includegraphics[width=\zisz]{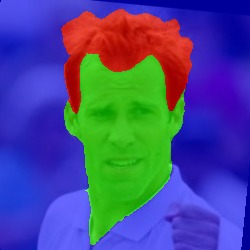}&
\includegraphics[width=\zisz]{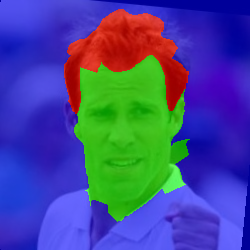}
&\includegraphics[width=\zisz]{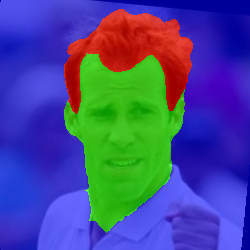}\\
&&$89.88\%$ 
&$98.05\%$\\
\includegraphics[width=\zisz]{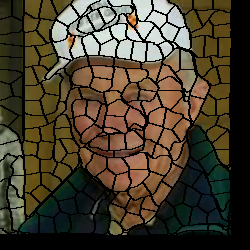}
&\includegraphics[width=\zisz]{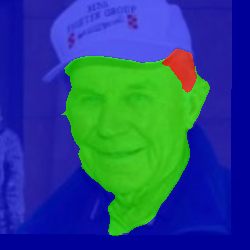}&
\includegraphics[width=\zisz]{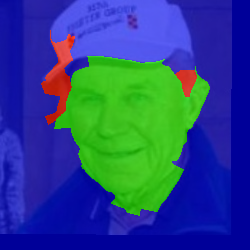}
&\includegraphics[width=\zisz]{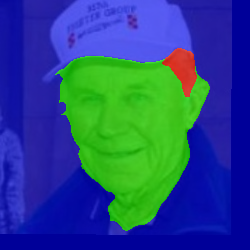}\\
Superpixels&Ground truth&SPM $R$=$0$&SPM $R$=$50$
\end{tabular}
\caption{
Labeling examples obtained with SPM
for $R$=$0$, 
and $R$=$50$ pixels, \emph{i.e.}, using superpatches,
with
superpixel-wise labeling accuracy.
}
\label{fig:LFW}
\end{scriptsize}
\end{figure*}

\begin{table}[t!]
\renewcommand{\arraystretch}{1.1}
\caption{Labeling accuracy on LFW
}
\centering
\newcommand{\sz}{\hspace{10pt}}
\newcommand{\szz}{\hspace{2pt}}
{\footnotesize
\begin{threeparttable}
\begin{tabular}{@{\szz}p{2.5cm}@{\sz}c@{\sz}c@{\sz}c@{\szz}}
\hline 
\multirow{2}{*}{Method}&Superpixel&Pixel&Computational \\
&accuracy&accuracy&time\\
\hline
{PatchMatch}& $87.73\%$&$87.02\%$&$3.940$s\\
Spatial CRF \cite{kae2013}&$93.95\%$ &\textit{not reported}& \textit{not reported}\\
CRBM \cite{kae2013}& $94.10\%$  &\textit{not reported}& \textit{not reported}\\
GLOC \cite{kae2013}& $94.95\%$  &\textit{not reported}& $0.323$s\\
DCNN \cite{liu2015}&\textit{not reported}& $95.24\%$& \textit{not reported} \\
SuperPatchMatch&$\mathbf{95.08\%}$ &$\mathbf{95.43\%}$&$\mathbf{0.255}$\textbf{s}\\
\hline 
\end{tabular}
\begin{tablenotes}
 \item Computational times are given per subject. 
SPM results are obtained with $k$=$50$ ANN, and $R$=$50$ pixels.
The presented values are the published results, therefore, some evaluation metrics could not be reported.
\end{tablenotes}
\end{threeparttable}
}
\label{table:LFW}
\end{table}

\subsubsection{Comparison with the state-of-the-art methods}
SPM is compared to the recent methods applied to the LFW
database in Table~\ref{table:LFW}. 
In \cite{kae2013}, the GLOC (GLObal and LOCal) method uses a restricted Boltzmann machine as complement of a  
conditional random field labeling \cite{lafferty2001}.
This combination reduces the error in face labeling of single models
which do not use global shape priors, at the expense of a higher computational cost. 
In \cite{liu2015}, a method based on a
deep convolutional neural network (DCNN) is proposed.
For all compared methods, learning steps, that can be up to several hours, are
necessary to train the models. 
Moreover, they consider priors learned from semantic information 
into the process, \emph{e.g.}, hair label should be on top of face label in the segmentation.
We also provide the results of a pixel-wise PM applied with the same framework,
where a SSD between patches of size $9{\times}9$ pixels in RGB color space is used as distance.

To compare to all methods, 
we provide in Table~\ref{table:LFW} superpixel and pixel-wise accuracy results.
The presented values 
are the results 
published by the authors, therefore, all the evaluation metrics could not be reported. 
SPM superpixel-wise labeling accuracy 
outperforms the ones of the compared methods ($95.08\%$), while being performed on basic features,
and faster ($0.255$s per subject) than the best compared method with reported computational time. 
The pixel-wise accuracy of SPM ($95.43\%$) also outperforms the reported result of the DCNN architecture \cite{liu2015},
that has been optimized to perform on the LFW dataset.
Note that the increase of pixel accuracy over superpixel accuracy
demonstrates that our method mostly fails at labeling small and stretched superpixels.
This comes from the initial LFW segmentation that may produce inaccurate color clustering
and allows irregular superpixel shapes.

The global computational time is another important comparison point.
SPM outperforms the compared methods in term of labeling accuracy without any training step.
Contrary to other methods,
with SPM, computational efforts needed for learning are canceled,
and new training images are directly considered in the library.
To illustrate this point, for each processed image, 
we add the remaining test ones to the library.
This way, SPM reaches $95.26\%$ of superpixel-wise labeling accuracy.
This result highlights the impact of 
the image diversity within the database, 
which leads to find more accurate ANN.
Moreover, results are obtained with no computational time increase, 
since the algorithm complexity only depends on the test image size. 
Hence, SPM easily integrates new images in the database, and 
provides very competitive results in limited computational time,
without model or shape \emph{priors}. \smallskip

\subsubsection{Robustness to superpixel decomposition method}

\begin{figure}[h!]
\centering
\begin{footnotesize}
\newcommand{\sizp}{0.235\columnwidth}
\newcommand{\sizpp}{0.2\columnwidth}
\begin{tabular}{@{\hspace{1mm}}c@{\hspace{1mm}}c@{\hspace{1mm}}c@{\hspace{1mm}}c@{\hspace{1mm}}}
\includegraphics[width=\sizp,height=\sizpp]{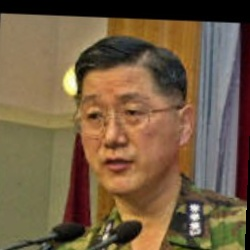}&
\includegraphics[width=\sizp,height=\sizpp]{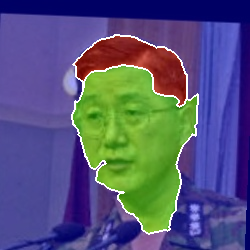}&
\includegraphics[width=\sizp,height=\sizpp]{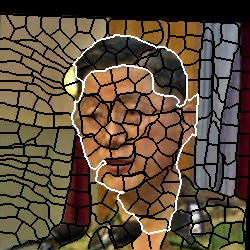}&
\includegraphics[width=\sizp,height=\sizpp]{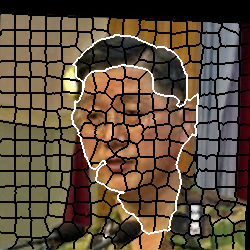}\\
(a)&(b)&(c)&(d)
\end{tabular}
\caption{Re-segmentation of the LFW dataset.
(a) Initial image. (b) Ground truth labels. (c) LFW initial decomposition. 
(d) Decomposition using \cite{achanta2012}.}
\label{fig:new_dec}
\end{footnotesize}
\end{figure}

To emphasize the robustness of our method to the used superpixel method,
we have segmented the test images with another method \cite{achanta2012}
that produces more regular superpixels (see an example in Fig. \ref{fig:new_dec}).
The new decompositions are computed with respect to the ground truth label mask of each image. 
Hence, they are still constrained by the initial segmentations provided with 
LFW but only on the edges of each class (hair, face, background).
Even with test and training decompositions computed with different methods,
we get similar superpixel-wise labeling accuracy ($95.05\%$), 
showing that our method can compare superpixel neighborhoods of various shapes.

\subsection{Non-Registered MRI Segmentation Experiments}

To demonstrate the robustness of the superpatch structure and 
the proposed framework,
we apply SPM to brain tumor segmentation on multi-modal non-registered 
Magnetic Resonance Images (MRI).
Classical patch-based and multi-atlas structure segmentation methods are 
based on registered subjects. Consequently, they cannot be efficiently applied 
in this non-registered context, due to the substantial variation in tumor shape and locations.
Superpixels enable to better capture the tumor geometry, thus
increasing the segmentation accuracy.
Superpixel and supervoxel-based approaches have been applied to tumor segmentation \cite{wang2013}. 
However, in this work, the neighborhood is not considered and the
ANN search is exhaustive, and computed on
a large multi-modal histogram descriptor, leading to prohibitive computational time.

SPM can be efficiently applied to tumor segmentation 
since it quickly finds good correspondences without image registration, 
and uses the superpixel neighborhood to improve the matching.
In this application, the segmentation is computed from a superpixel decomposition \cite{achanta2012},
then each region (tumor or background) is labeled with SPM.

We present results obtained on the MICCAI multi-modal Brain Tumor
Segmentation (BRATS) dataset \cite{brats2012}.
This challenging dataset contains real and simulated patient data, 
with overall poor resolution and large variation of tumor shape and position.
For both types, high grade (HG) and low grade (LG) tumors are provided with four modalities: 
T1, contrast enhanced T1 (T1C), T2, and FLAIR. 
Overall, there are 20 and 10 real patient data with respectively HG and LG tumors,
and 25 images for both HG and LG simulated tumor data.
We use the same SPM parameters as in Section \ref{section:xps}, 
taking a
multi-modal histogram, containing the levels of gray intensity on all MRI modalities
as descriptor for superpatch matching,
and performing the regularization \eqref{gc} at the pixel scale to compare with pixel-wise ground truths.
Each subject is segmented by the remaining of its type in a leave-one-out procedure.

 \begin{table}[t]
\caption{Dice coefficient and computational time results for different structure descriptors}
\centering
\newcommand{\sz}{\hspace{6pt}}
\newcommand{\szz}{\hspace{1pt}}
\newcommand{\szzz}{\hspace{2pt}}
\begin{footnotesize}
\begin{tabular}{@{\szz}p{2cm}@{\sz}c@{\sz}c@{\sz}c@{\sz}c@{\szzz}c@{\szz}}
\hline
\multirow{2}{*}{Method}&\multicolumn{2}{c}{\hspace{-0.3cm}Simulated Data}&\multicolumn{2}{c}{\hspace{-0.3cm}Real Data}&{Computational}\\
&HG&LG&HG&LG& time\\
\hline
Superpixel-based &$73.95\%$&$41.95\%$&$44.63\%$&$38.07\%$&$\mathbf{0.33}$\textbf{s}\\
Patch-based &$69.11\%$&$49.55\%$&$52.59\%$&$68.92\%$&$2.69$s\\
Superpatch-based&$\mathbf{90.75\%}$&$\mathbf{82.40\%}$&$\mathbf{61.87\%}$&$\mathbf{73.34\%}$&$0.94$s\\
\hline  
\end{tabular}
\end{footnotesize}
\label{table:brats_table}
\end{table}

\newcommand\ziszz{0.12\textwidth}
\newcommand\hhh{60pt}
\newcommand\ziszzk{0.085\textwidth}
\newcommand\hhhk{40pt}

\begin{figure*}[ht]
{\scriptsize
\centering 
\begin{tabular}{@{\hspace{0mm}}l@{\hspace{1mm}}c@{\hspace{1mm}}c@{\hspace{1mm}}c@{\hspace{1mm}}c@{\hspace{2mm}}}   
\multirow{4}{*}{\rotatebox{90}{\parbox{0.15\textwidth}{Simulated HG} \hspace{-1.6cm} }}&
\includegraphics[width=\ziszz,height=\hhh]{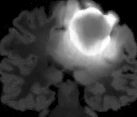}&
\includegraphics[width=\ziszz,height=\hhh]{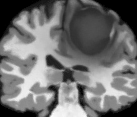}&
\includegraphics[width=\ziszz,height=\hhh]{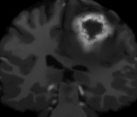}&
\includegraphics[width=\ziszz,height=\hhh]{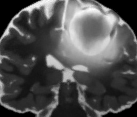}\\
&FLAIR&T1&T1C&T2\\
&\includegraphics[width=\ziszz,height=\hhh]{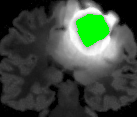}&
\includegraphics[width=\ziszz,height=\hhh]{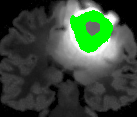}&
\includegraphics[width=\ziszz,height=\hhh]{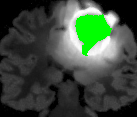}&
\includegraphics[width=\ziszz,height=\hhh]{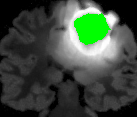}\\ 
&Ground truth&Patch-based &SPM $R$=$0$&SPM $R$=$25$\\
\end{tabular} 
\begin{tabular}{@{\hspace{0mm}}l@{\hspace{1mm}}c@{\hspace{1mm}}c@{\hspace{1mm}}c@{\hspace{1mm}}c@{\hspace{1mm}}c@{\hspace{0mm}}} 
{\rotatebox{90}{\parbox{0.08\textwidth}{ Simulated LG} \hspace{-1.5cm} }}&
\includegraphics[width=\ziszzk,height=\hhhk]{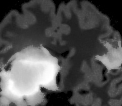}&
\includegraphics[width=\ziszzk,height=\hhhk]{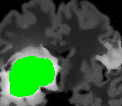}&
\includegraphics[width=\ziszzk,height=\hhhk]{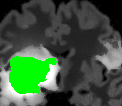}&
\includegraphics[width=\ziszzk,height=\hhhk]{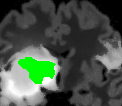}&
\includegraphics[width=\ziszzk,height=\hhhk]{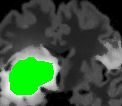}\\
{\rotatebox{90}{\parbox{0.08\textwidth}{\hspace{0.2cm} Real HG}  }}&
\includegraphics[width=\ziszzk,height=\hhhk]{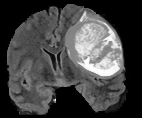}&
\includegraphics[width=\ziszzk,height=\hhhk]{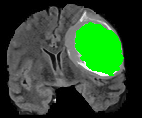}&
\includegraphics[width=\ziszzk,height=\hhhk]{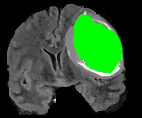}&
\includegraphics[width=\ziszzk,height=\hhhk]{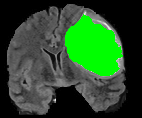}&
\includegraphics[width=\ziszzk,height=\hhhk]{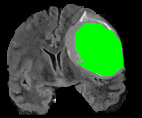}\\
{\rotatebox{90}{\parbox{0.08\textwidth}{\hspace{0.3cm}Real LG} }}&
\includegraphics[width=\ziszzk,height=\hhhk]{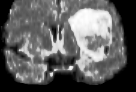}&
\includegraphics[width=\ziszzk,height=\hhhk]{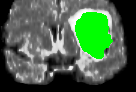}&
\includegraphics[width=\ziszzk,height=\hhhk]{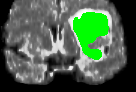}&
\includegraphics[width=\ziszzk,height=\hhhk]{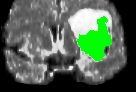}&
\includegraphics[width=\ziszzk,height=\hhhk]{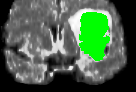}\\
&FLAIR&Ground truth&Patch-based &SPM $R$=$0$&SPM $R$=$25$\\
\end{tabular}
\caption{
Examples of patch, superpixel and superpatch-based tumor segmentation results.
The four modalities are displayed for simulated HG data.} 
\label{fig:BRATS}
}
\end{figure*}

In Fig. \ref{fig:BRATS}, we show several tumor segmentation results for all data types.
In Table \ref{table:brats_table}, we compare results obtained using different descriptor structures:
patch-based \cite{giraud2016}, superpixel-based \cite{wang2013},
and superpatch-based ($R$=$25$ pixels).
We use the Dice coefficient \cite{zijdenbos1994} as evaluation metric,
measuring the overlap between the automatically segmented structure and the ground truth. 
The superpixel-based approach
appears very limited since it fails at capturing the tumor context and their location in other images.
Regular patches are also limited in this context, due to the variations in the structure shapes.
Superpatches provide a robust descriptor, since they follow image intensities and capture the superpixel neighborhood, 
leading to more accurate segmentation.
These experiments demonstrate that superpatches within the SPM framework provide fast and accurate segmentation results
even on non-registered multi-modal images with poor resolution. \\

\section{Conclusion and Perspectives}
In this paper, we propose a new structure based on patches of superpixels  
that can use irregular and non stable image  
decompositions. 
These superpatches include neighborhood information and lead to more accurate matching. 
We also introduce SuperPatchMatch, a general and novel
correspondence algorithm
of superpatches.

We have demonstrated the interest of our framework by obtaining state-of-the-art results
for face labeling and tumor segmentation on non-registered MRI.
SuperPatchMatch does not need any learning phase, 
that can be up to several hours for many methods of the literature.
By including spatial consistency,
superpatches are able to reach the accuracy of highly tuned approaches,
and provide more reliable descriptors than single superpixels. %

Our work opens new insights for future adaptations to superpixel-based methods, 
\emph{e.g.}, 
segmentation \cite{fulkerson2009,trulls2014}, 
labeling \cite{gould2014},
saliency detection \cite{pei2014}, or
 color and style transfer \cite{liu2016photo}.
For instance, SuperPatchMatch can be considered for defining good ANN 
initializations at the pixel level, when the size of the database is too large.
A possible application is the  optical flow initialization, instead of mutli-resolution schemes, 
to better capture large displacements of small objects.

\bibliographystyle{IEEEtran}
\bibliography{TIP_SPM}

\begin{thebibliography}{10}
\providecommand{\url}[1]{#1}
\csname url@samestyle\endcsname
\providecommand{\newblock}{\relax}
\providecommand{\bibinfo}[2]{#2}
\providecommand{\BIBentrySTDinterwordspacing}{\spaceskip=0pt\relax}
\providecommand{\BIBentryALTinterwordstretchfactor}{4}
\providecommand{\BIBentryALTinterwordspacing}{\spaceskip=\fontdimen2\font plus
\BIBentryALTinterwordstretchfactor\fontdimen3\font minus
  \fontdimen4\font\relax}
\providecommand{\BIBforeignlanguage}[2]{{%
\expandafter\ifx\csname l@#1\endcsname\relax
\typeout{** WARNING: IEEEtran.bst: No hyphenation pattern has been}%
\typeout{** loaded for the language `#1'. Using the pattern for}%
\typeout{** the default language instead.}%
\else
\language=\csname l@#1\endcsname
\fi
#2}}
\providecommand{\BIBdecl}{\relax}
\BIBdecl

\bibitem{achanta2012}
R.~Achanta, A.~Shaji, K.~Smith, A.~Lucchi, P.~Fua, and S.~Süsstrunk, ``{SLIC}
  superpixels compared to state-of-the-art superpixel methods,'' \emph{IEEE
  Trans. Pattern Anal. Mach. Intell.}, vol.~34, no.~11, pp. 2274--2282, 2012.

\bibitem{barnes2009}
C.~Barnes, E.~Shechtman, A.~Finkelstein, and D.~Goldman, ``Patch{M}atch: A
  randomized correspondence algorithm for structural image editing,'' \emph{ACM
  Trans. Graph.}, vol.~28, no.~3, 2009.

\bibitem{muja2009fast}
M.~Muja and D.~G. Lowe, ``Fast approximate nearest neighbors with automatic
  algorithm configuration,'' vol.~2, 2009, pp. 331--340.

\bibitem{korman2011}
S.~Korman and S.~Avidan, ``Coherency sensitive hashing,'' in \emph{Proc. IEEE
  ICCV}, 2011, pp. 1607--1614.

\bibitem{olonetsky2012}
I.~Olonetsky and S.~Avidan, ``{TreeCANN} - k-d tree coherence approximate
  nearest neighbor algorithm,'' in \emph{Proc. ECCV}, 2012, pp. 602--615.

\bibitem{barnes2010}
C.~Barnes, E.~Shechtman, D.~B. Goldman, and A.~Finkelstein, ``The generalized
  {PatchMatch} correspondence algorithm,'' in \emph{Proc. ECCV}, 2010, pp.
  29--43.

\bibitem{wang2011}
S.~Wang, H.~Lu, F.~Yang, and M.~H. Yang, ``Superpixel tracking,'' in
  \emph{Proc. IEEE ICCV}, 2011, pp. 1323--1330.

\bibitem{reso2013}
M.~Reso, J.~Jachalsky, B.~Rosenhahn, and J.~Ostermann, ``Temporally consistent
  superpixels,'' in \emph{Proc. IEEE ICCV}, 2013, pp. 385--392.

\bibitem{Rabin_icip14}
J.~Rabin, S.~Ferradans, and N.~Papadakis, ``Adaptive color transfer with
  relaxed optimal transport,'' in \emph{Proc. IEEE ICIP}, 2014, pp. 4852--4856.

\bibitem{liu2016photo}
J.~Liu, W.~Yang, X.~Sun, and W.~Zeng, ``Photo stylistic brush: Robust style
  transfer via superpixel-based bipartite graph,'' \emph{arXiv preprint
  arXiv:1606.03871}, 2016.

\bibitem{pei2014}
S.-C. Pei, W.-W. Chang, and C.-T. Shen, ``Saliency detection using superpixel
  belief propagation,'' in \emph{Proc. IEEE ICIP}, 2014, pp. 1135--1139.

\bibitem{sawhney2014}
R.~Sawhney, F.~Li, and H.~I. Christensen, ``{GASP}: Geometric association with
  surface patches,'' in \emph{Proc. 3DV}, 2014, pp. 107--114.

\bibitem{buyssens2014}
P.~Buyssens, M.~Toutain, A.~Elmoataz, and O.~L{\'e}zoray, ``{Eikonal-based
  vertices growing and iterative seeding for efficient graph-based
  segmentation},'' in \emph{{Proc. IEEE ICIP}}, 2014, pp. 4368--4372.

\bibitem{gould2014}
S.~Gould, J.~Zhao, X.~He, and Y.~Zhang, ``Superpixel graph label transfer with
  learned distance metric,'' in \emph{Proc. ECCV}, 2014, pp. 632--647.

\bibitem{zheng2015}
J.~Zheng and Z.~Li, ``Superpixel based patch match for differently exposed
  images with moving objects and camera movements,'' in \emph{IEEE ICIP}, 2015,
  pp. 4516--4520.

\bibitem{lu2013}
J.~Lu, H.~Yang, D.~Min, and M.~N. Do, ``Patch{M}atch filter: Efficient
  edge-aware filtering meets randomized search for fast correspondence field
  estimation,'' in \emph{Proc. IEEE CVPR}, 2013, pp. 1854--1861.

\bibitem{he2004}
X.~He, R.~Zemel, and M.~Carreira-Perpi{\~n}{\'a}n, ``Multiscale conditional
  random fields for image labeling,'' in \emph{Proc. IEEE CVPR}, vol.~2, 2004.

\bibitem{kae2013}
A.~Kae, K.~Sohn, H.~Lee, and E.~Learned-Miller, ``Augmenting {CRF}s with
  {B}oltzmann machine shape priors for image labeling,'' in \emph{Proc. IEEE
  CVPR}, 2013, pp. 2019--2026.

\bibitem{liu2015}
S.~Liu, J.~Yang, C.~Huang, and M.~Yang, ``Multi-objective convolutional
  learning for face labeling,'' in \emph{Proc. IEEE CVPR}, 2015, pp.
  3451--3459.

\bibitem{long2015}
J.~Long, E.~Shelhamer, and T.~Darrell, ``Fully convolutional networks for
  semantic segmentation,'' in \emph{Proc. IEEE CVPR}, 2015, pp. 3431--3440.

\bibitem{huang2007db}
G.~Huang, M.~Ramesh, T.~Berg, and E.~Learned-Miller, ``Labeled faces in the
  wild: A database for studying face recognition in unconstrained
  environments,'' \emph{Tech. Rep. 07-49, Univ. of Massachusetts, Amherst},
  vol.~1, no.~2, 2007.

\bibitem{dalal2005}
N.~Dalal and B.~Triggs, ``Histograms of oriented gradients for human
  detection,'' in \emph{Proc. IEEE CVPR}, 2005, pp. 886--893.

\bibitem{vincent1991}
L.~Vincent and P.~Soille, ``Watersheds in digital spaces: An efficient
  algorithm based on immersion simulations,'' \emph{IEEE Trans. Pattern Anal.
  Mach. Intell.}, vol.~13, no.~6, pp. 583--598, 1991.

\bibitem{vedaldi2008}
A.~Vedaldi and S.~Soatto, ``Quick shift and kernel methods for mode seeking,''
  in \emph{Proc. ECCV}, 2008, pp. 705--718.

\bibitem{li2015}
Z.~Li and J.~Chen, ``Superpixel segmentation using linear spectral
  clustering,'' in \emph{Proc. IEEE CVPR}, 2015, pp. 1356--1363.

\bibitem{machairas2015}
V.~Machairas, M.~Faessel, D.~C{\'a}rdenas-Pe{\~n}a, T.~Chabardes, T.~Walter,
  and E.~Decenci{\`e}re, ``{Waterpixels},'' \emph{{IEEE Trans. Image
  Process.}}, vol.~24, no.~11, pp. 3707--3716, 2015.

\bibitem{giraud2016scalp}
R.~Giraud, V.-T. Ta, and N.~Papadakis, ``{SCALP: S}uperpixels with contour
  adherence using linear path,'' in \emph{{Proc. ICPR}}, 2016, pp. 2374--2379.

\bibitem{ban2016}
Z.~Ban, J.~Liu, and J.~Fouriaux, ``{GLSC: LSC superpixels at over 130 FPS},''
  \emph{J. Real-Time Image Process.}, pp. 1--12, 2016.

\bibitem{gould2008}
S.~Gould, J.~Rodgers, D.~Cohen, G.~Elidan, and D.~Koller, ``Multi-class
  segmentation with relative location prior,'' \emph{Int. J. Comput. Vis.},
  vol.~80, no.~3, pp. 300--316, 2008.

\bibitem{tighe2010}
J.~Tighe and S.~Lazebnik, ``{SuperParsing}: Scalable nonparametric image
  parsing with superpixels,'' in \emph{Proc. ECCV}, 2010, pp. 352--365.

\bibitem{yang2010}
Y.~Yang, S.~Hallman, D.~Ramanan, and C.~Fowlkes, ``Layered object detection for
  multi-class segmentation,'' in \emph{Proc. IEEE CVPR}, 2010, pp. 3113--3120.

\bibitem{mostajabi2015}
M.~Mostajabi, P.~Yadollahpour, and G.~Shakhnarovich, ``Feedforward semantic
  segmentation with zoom-out features,'' in \emph{Proc. IEEE CVPR}, 2015, pp.
  3376--3385.

\bibitem{mori2005}
G.~Mori, ``Guiding model search using segmentation,'' in \emph{Proc. IEEE
  ICCV}, 2005, pp. 1417--1423.

\bibitem{fulkerson2009}
B.~Fulkerson, A.~Vedaldi, and S.~Soatto, ``Class segmentation and object
  localization with superpixel neighborhoods,'' in \emph{Proc. IEEE ICCV},
  2009, pp. 670--677.

\bibitem{arbelaez2011}
P.~Arbeláez, M.~Maire, C.~Fowlkes, and J.~Malik, ``Contour detection and
  hierarchical image segmentation,'' \emph{IEEE Trans. Pattern Anal. Mach.
  Intell.}, vol.~33, no.~5, pp. 898--916, 2011.

\bibitem{Efros99}
A.~Efros and T.~Leung, ``Texture synthesis by non-parametric sampling,'' in
  \emph{Proc. IEEE ICCV}, 1999, pp. 1033--1038.

\bibitem{buades2005}
A.~Buades, B.~Coll, and J.-M. Morel, ``A non-local algorithm for image
  denoising,'' in \emph{Proc. IEEE CVPR}, 2005, pp. 60--65.

\bibitem{Lowe2004}
D.~G. Lowe, ``Distinctive image features from scale-invariant keypoints,''
  \emph{Int. J. Comput. Vis.}, vol.~60, no.~2, pp. 91--110, 2004.

\bibitem{bay2006}
H.~Bay, T.~Tuytelaars, and L.~Van~Gool, ``{SURF: S}peeded up robust features,''
  in \emph{Proc. ECCV}, 2006, pp. 404--417.

\bibitem{garnier2012}
M.~Garnier, T.~Hurtut, and L.~Wendling, ``Object description based on spatial
  relations between level-sets,'' in \emph{Proc. DICTA}, 2012, pp. 1--7.

\bibitem{clement2015}
M.~Cl{\'e}ment, M.~Garnier, C.~Kurtz, and L.~Wendling, ``Color object
  recognition based on spatial relations between image layers,'' in
  \emph{{Proc. VISAPP}}, 2015, pp. 427--434.

\bibitem{lsvm-pami}
P.~F. Felzenszwalb, R.~B. Girshick, D.~McAllester, and D.~Ramanan, ``Object
  detection with discriminatively trained part based models,'' \emph{IEEE
  Trans. Pattern Anal. Mach. Intell.}, vol.~32, no.~9, pp. 1627--1645, 2010.

\bibitem{trulls2014}
E.~Trulls, S.~Tsogkas, I.~Kokkinos, A.~Sanfeliu, and F.~Moreno-Noguer,
  ``Segmentation-aware deformable part models,'' in \emph{Proc. IEEE CVPR},
  2014, pp. 168--175.

\bibitem{sharma2013}
G.~Sharma, F.~Jurie, and C.~Schmid, ``Expanded parts model for human attribute
  and action recognition in still images,'' in \emph{Proc. IEEE CVPR}, 2013,
  pp. 652--659.

\bibitem{bloch2005}
I.~Bloch, ``Fuzzy spatial relationships for image processing and
  interpretation: A review,'' \emph{Image and Vision Comp.}, vol.~23, no.~2,
  pp. 89--110, 2005.

\bibitem{Freeman02}
W.~Freeman, T.~Jones, and E.~Pasztor, ``Example-based super-resolution,''
  \emph{IEEE Trans. Comp. Graph. App.}, vol.~22, no.~2, pp. 56--65, 2002.

\bibitem{shi2013}
W.~Shi, J.~Caballero, C.~Ledig, X.~Zuang, W.~Bai, K.~Bhatia, A.~Marvao,
  T.~Dawes, D.~O'Regan, and D.~Rueckert, ``Cardiac image super-resolution with
  global correspondence using multi-atlas {PatchMatch},'' in \emph{Proc.
  MICCAI}, 2013, pp. 9--16.

\bibitem{giraud2016}
R.~Giraud, V.-T. Ta, N.~Papadakis, J.~V. Manj{\'o}n, D.~L. Collins,
  P.~Coup{\'e}, and the Alzheimer’s Disease Neuroimaging~Initiative, ``An
  optimized {PatchMatch} for multi-scale and multi-feature label fusion,''
  \emph{NeuroImage}, vol. 124, pp. 770--782, 2016.

\bibitem{brats2012}
B.~H. Menze, A.~Jakab, S.~Bauer, J.~Kalpathy-Cramer, K.~Farahani, J.~Kirby,
  Y.~Burren, N.~Porz, J.~Slotboom, R.~Wiest \emph{et~al.}, ``The multimodal
  brain tumor image segmentation benchmark {(BRATS)},'' \emph{{IEEE Trans. Med.
  Imaging}}, vol.~34, no.~10, pp. 1993--2024, 2015.

\bibitem{coupe2011patch}
P.~Coup{\'e}, J.~V. Manj{\'o}n, V.~Fonov, J.~Pruessner, M.~Robles, and D.~L.
  Collins, ``Patch-based segmentation using expert priors: application to
  hippocampus and ventricle segmentation,'' \emph{NeuroImage}, vol.~54, no.~2,
  pp. 940--954, 2011.

\bibitem{huang2007}
G.~Huang, V.~Jain, and E.~Learned-Miller, ``Unsupervised joint alignment of
  complex images,'' in \emph{Proc. IEEE ICCV}, 2007, pp. 1--8.

\bibitem{boykov2001}
Y.~Boykov, O.~Veksler, and R.~Zabih, ``Fast approximate energy minimization via
  graph cuts,'' \emph{IEEE Trans. Pattern Anal. Mach. Intell.}, vol.~23,
  no.~11, pp. 1222--1239, 2001.

\bibitem{lafferty2001}
J.~Lafferty, A.~McCallum, and F.~Pereira, ``Conditional random fields:
  Probabilistic models for segmenting and labeling sequence data,'' in
  \emph{Proc. ICML}, 2001, pp. 282--289.

\bibitem{wang2013}
H.~Wang and P.~A. Yushkevich, ``Multi-atlas segmentation without registration:
  A supervoxel-based approach,'' in \emph{Proc. MICCAI}, 2013, pp. 535--542.

\bibitem{zijdenbos1994}
A.~Zijdenbos, B.~Dawant, R.~Margolin, and A.~Palmer, ``Morphometric analysis of
  white matter lesions in {MR} images: method and validation,'' \emph{IEEE
  Trans. Med. Imaging}, vol.~13, no.~4, pp. 716--724, 1994.

\end{thebibliography}

\vspace{-0.5cm}
\begin{IEEEbiography}[{\includegraphics[width=1in,height=1.25in,clip,keepaspectratio]{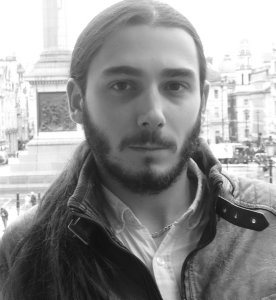}}]%
{Rémi Giraud}
received the M.Sc. in telecommunications at ENSEIRB-MATMECA School of Engineers, 
and the M.Sc. in signal and image processing from the University of Bordeaux, France, in 2014.
Since, he is pursuing his Ph.D. at the Laboratoire  Bordelais  de  Recherche   en
Informatique
in the field of image processing.
His research areas mainly include computer vision and image processing applications
with patch-based and superpixel methods.
\end{IEEEbiography}\vspace{-1cm}
\begin{IEEEbiography}[{\includegraphics[width=1in,height=1.25in,clip,keepaspectratio]{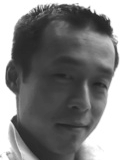}}]%
{Vinh-Thong Ta}
received the M.Sc. and Doctoral degrees in computer science from the University of Caen Basse-Normandie, France, in 2004
and 2009, respectively. From 2009 to 2010, he was an Assistant Professor in computer science
with the School of Engineers of Caen, France.
Since 2010, he is an Associate Professor with the Computer Science Department, School of Engineers
ENSEIRB-MATMECA. His research mainly concerns image and
data processing. 
\end{IEEEbiography}\vspace{-1cm}
\begin{IEEEbiography}[{\includegraphics[width=1in,height=1.25in,clip,keepaspectratio]{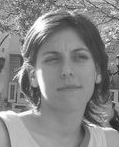}}]%
{Aurélie Bugeau}
received the Ph.D. degree in signal
processing  from  the  University  of  Rennes,  France,
in 2007.  She  is  an  Associate  Professor  with  the
University of Bordeaux and was involved in research
with  the  Laboratoire   Bordelais   de  Recherche   en
Informatique.  Her  main  research  interests  include
patch-based and graph-based methods for image and
video  processing,  inpainting,  and editing.
\end{IEEEbiography}\vspace{-1cm}
\begin{IEEEbiography}[{\includegraphics[width=1in,height=1.25in,clip,keepaspectratio]{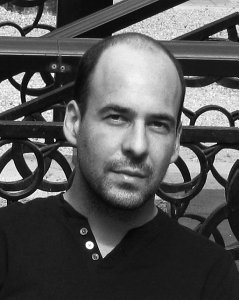}}]%
{Pierrick Coupé}
received a M. Eng degree in Biomedical
engineering from the University of Technology of Compiègne in 2003,
a M.Sc in Image and Signal processing degree from
University of Rennes I in 2004, and obtained a Ph.D. degree in
Image processing from the University of Rennes I in 2008.
Since 2011, he is a full time CNRS researcher at the
Laboratoire Bordelais de Recherche en Informatique.
His research mainly focuses on medical image enhancement 
and analysis, and computer-aided diagnosis.
\end{IEEEbiography}\vspace{-1cm}
\begin{IEEEbiography}[{\includegraphics[width=1in,height=1.25in,clip,keepaspectratio]{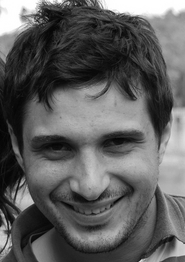}}]%
{Nicolas Papadakis}
received the Graduate degree in
applied mathematics from the National Institute of Applied Sciences, Rouen, France, in 2004, and the 
Ph.D. degree in applied mathematics from the University of Rennes, France, in 2007. 
He is currently a  Researcher from the Centre National de la Recherche Scientifique, Institut de Mathématiques
de Bordeaux, France. His main research interests include tracking, motion estimation, and optimal
transportation for image processing.
\end{IEEEbiography}

\begin{sloppypar}

\end{sloppypar}

\end{document}